\documentclass{article}
 \usepackage[preprint]{log_2026}			% for preprint version

\usepackage[utf8]{inputenc}
\usepackage[T1]{fontenc}
\usepackage{url}

\usepackage{nicefrac}
\usepackage{microtype}
\usepackage{xcolor}
\usepackage{amsmath}
\usepackage{amssymb}
\usepackage{wrapfig}

\usepackage{booktabs}						% professional-quality tables
\usepackage{multirow}						% tabular cells spanning multiple rows
\usepackage{amsfonts}						% blackboard math symbols
\usepackage{graphicx}						% figures
\usepackage{duckuments}						% sample images
\usepackage{longtable}
\usepackage{caption}

\usepackage[numbers,compress,sort]{natbib}	% for numerical citations
\title{Beyond Fixed Features: Architecture-Dependent Sensitivity to Node Representations under Heterophily}

\author{%
Priyanath Maji\\
Georgia Institute of Technology \\
\email{pmaji3@gatech.edu}\And
Sidharth Gaur\\
Georgia Institute of Technology\\
\email{sgaur38@gatech.edu}\And
Rajavinoth Paul Durai\\
Georgia Institute of Technology\\
\email{rdurai6@gatech.edu}
}

\begin{document}

\maketitle

\begin{abstract}
Graph Neural Networks (GNNs) perform well on homophilic graphs but struggle in heterophilic settings, where connected nodes often carry dissimilar labels. Existing evaluations typically compare architectures under a fixed node-feature representation, leaving unclear whether conclusions about heterophily robustness remain stable as the input representation changes. We address this question by constructing parallel feature variants of two large-scale heterophilic benchmarks, Roman-Empire and Amazon-Ratings, pairing each graph with representations ranging from static fastText vectors to contextual Transformer embeddings and evaluating seven GNN architectures across these representations. We find that the effect of representation varies across architectures: on Roman-Empire, the contextual gain ranges from 2.38 percentage points for GCN-sep to 13.67 points for GAT, with H2GCN gaining 8.77 points. On Amazon-Ratings, where node text is limited to short product titles, GAT improves by 6.78 points from fastText to MPNet, while GCN-sep changes by only 0.20 points. These results show that architectural performance is conditional on node representation: the same representation change can produce different magnitudes of performance gain across architectures, so architecture and representation cannot be treated as independent evaluation factors. A rank-correlation analysis on these two benchmarks further shows that the relative ordering of architectures remains highly stable across representations, isolating differential sensitivity, rather than ranking instability, as the primary effect.
\end{abstract}

%%%%%%%%% BODY TEXT

\section{Introduction}

Graph Neural Networks (GNNs) have become a standard paradigm for graph representation learning, with applications spanning social network analysis, recommendation systems, and molecular property prediction. At their core, GNNs use a spatial message-passing framework in which node representations are updated by aggregating information from local neighborhoods. This mechanism is highly effective on \emph{homophilic} graphs, where connected nodes tend to exhibit similar labels, but becomes more challenging in \emph{heterophilic} settings such as fraud detection and web networks, where adjacency does not imply class similarity. When neighboring nodes carry dissimilar information, neighborhood aggregation can mix useful node-specific signal with information that is less relevant to the prediction task. This has motivated a broad range of architectural approaches for heterophilic graphs, including ego-neighbor separation in H2GCN~\cite{H2GCN}, initial residual and identity mappings in GCNII~\cite{chen2020simpledeepgraphconvolutionalGCNII}, and adaptive aggregation mechanisms such as GPRGNN~\cite{GPRGNN}, FSGNN~\cite{FSGNN}, and GloGNN~\cite{GLOGNN}. Platonov et al.~\cite{heterophily} further demonstrated that reliable evaluation of heterophilic GNNs depends strongly on benchmark quality and introduced Roman-Empire and Amazon-Ratings as larger and more reliable alternatives to several widely used small-scale benchmarks. While this work has substantially advanced our understanding of architectural behavior under heterophily, most architecture comparisons still evaluate models using a single, fixed node-feature representation. This leaves an important question open: \emph{does the relative behavior of GNN architectures depend on the representation provided at their input?}

This question is important because node features contain information that graph propagation cannot necessarily recover. A fixed feature representation can therefore become an implicit part of an architecture comparison, even when the stated goal is to compare architectures themselves. For example, in Roman-Empire, node labels correspond to syntactic roles in a dependency graph. The predictive information associated with a word can therefore depend on its surrounding sentence and its relation to other words, rather than solely on its lexical identity. A static embedding assigns the same representation to repeated occurrences of a word, whereas a contextual Transformer representation can encode occurrence-specific information from the surrounding context. We therefore expect contextualized representations to provide greater task-relevant information than static lexical representations on Roman-Empire. On Amazon-Ratings, where nodes consist primarily of short product titles, contextualization may provide a smaller and qualitatively different benefit by capturing semantic relationships between differently worded titles.

More broadly, we ask whether improvements in node representation affect all GNN architectures similarly or whether their benefits depend on the architecture that processes them. One possibility is that stronger representations compensate for architectural limitations, causing performance differences between architectures to shrink as the input representation becomes more informative. Another is that representation and architecture contribute complementary information, such that improving the representation produces different gains for different architectures. Distinguishing these possibilities requires varying the representation while holding the underlying graph fixed rather than comparing architectures across independently chosen feature settings.

We study this question on Roman-Empire (22,600 nodes) and Amazon-Ratings (24,492 nodes), whose adjusted homophily values are $-0.055$ and $0.141$, respectively. For each graph, we construct multiple node-feature representations ranging from static fastText~\cite{grave2018learningwordvectors157fastText} vectors to contextual Transformer embeddings, including BERT-base-uncased~\cite{devlin2018bert_bert-base-uncased}, RoBERTa~\cite{liu2019robertarobustlyoptimizedbert_RoBERTa}, Sentence-BERT (SBERT)~\cite{reimers-2019-sentence-bert_SBERT}, and MPNet~\cite{MPNET}, while keeping the graph structure unchanged. We evaluate seven GNN architectures and an MLP control across all representation-architecture combinations using a 10-fold cross-validation protocol. The MLP provides a feature-only reference, allowing us to assess performance obtainable from the node representation without graph propagation. We also include GCN-sep and GAT-sep as architecture variants motivated by established heterophily approaches, without treating the separation mechanism itself as a novel contribution.

Our fastText setting reproduces the published architecture ordering reported by Platonov et al.~\cite{heterophily}, providing a consistency check on our benchmark reconstruction. More importantly, varying the node representation reveals systematic differences that are not visible in a single-feature evaluation. Contextualized representations improve the performance of the strongest architectures on both benchmarks: moving from static to contextual embeddings improves GAT-sep by 4.24 percentage points on Roman-Empire and GraphSAGE~\cite{graphSAGE} by 1.46 points on Amazon-Ratings for the same architecture. The magnitude of the improvement, however, varies across architectures, indicating that the effect of changing the node representation is architecture-dependent rather than uniform. These findings show that node representation and architectural design cannot be cleanly evaluated as independent factors: conclusions about which architecture performs best under heterophily can depend on the feature representation supplied to the model. Evaluations that hold node features fixed may therefore overlook an important source of variation in GNN performance.

\section{Related Work}

\noindent\textbf{Heterophily and Message Passing.}
Standard GNNs such as GCN~\cite{GCN} and GAT~\cite{GAT} derive node representations from local neighborhood information, an approach that becomes less effective when neighboring nodes exhibit weak or negative label correlation. This has motivated architectural approaches specifically designed for heterophilic graphs. H2GCN~\cite{H2GCN} separates ego-node and neighborhood representations and incorporates multi-hop information, while GCNII~\cite{chen2020simpledeepgraphconvolutionalGCNII} uses initial residual connections and identity mapping to improve information preservation in deeper networks. Other approaches, including GPR-GNN, FSGNN, and GloGNN, use adaptive or learned aggregation mechanisms to address limitations of conventional message passing. These studies establish that architectural design is an important component of GNN performance under heterophily. In our experiments, we include several established architectures, including separation-based variants, as part of a broader comparison across feature representations; we do not claim the underlying architectural mechanisms as new.

\noindent\textbf{Benchmark Reliability.}
Platonov et al.~\cite{heterophily} examined the reliability of commonly used heterophily benchmarks and showed that small datasets and high variance can produce unstable or misleading architectural comparisons. They introduced Roman-Empire and Amazon-Ratings as larger benchmarks with substantially more reliable evaluation properties and proposed adjusted homophily as a measure of label agreement relative to a degree-weighted null model. We build on these benchmarks and reconstruct the datasets from their primary sources. Unlike prior architecture comparisons that typically use a fixed feature representation for each benchmark, our evaluation varies the node representation while keeping the underlying graph unchanged. Adjusted homophily is used here to characterize the benchmark graphs rather than as a direct measure of neighborhood informativeness.

\noindent\textbf{Node Representations in Graph Learning.}
Node features are a fundamental component of graph learning, but their role is often treated as fixed when evaluating GNN architectures. In text-attributed graphs, representations can range from static lexical embeddings, which assign a context-independent representation to each token, to contextualized Transformer embeddings, which incorporate information from the surrounding sequence. The resulting representations differ in the type and amount of information they encode, but they need not form a universal quality ordering. We therefore avoid treating embedding families themselves as a fixed ranking of feature quality. Instead, we evaluate their task-relevant informativeness empirically using an MLP control and then examine how the same representations perform across different GNN architectures. This allows us to study whether improvements in the information available at the node level translate uniformly across architectures or interact with the way each architecture uses graph structure.

\section{Rationale and Hypothesis}
\label{sec:rationale}

Our first hypothesis concerns the information encoded by the node representation. We hypothesize that contextualized embeddings provide greater task-relevant information than static lexical embeddings when prediction depends on contextual information. This distinction is particularly relevant to Roman-Empire, where node labels represent syntactic roles in a dependency graph. The same lexical item can occur in different syntactic roles depending on its surrounding sentence and its structural relation to other words. A static embedding cannot distinguish these occurrences, whereas a contextual representation can encode information specific to each occurrence. We therefore expect contextualized Transformer embeddings to improve task performance relative to fastText on Roman-Empire. We expect the effect to be smaller on Amazon-Ratings, where node text consists primarily of short product titles and contextual information is less directly tied to syntactic role.

Our second hypothesis concerns the relationship between node representation and architecture. Existing GNN evaluations generally compare architectures while holding the node representation fixed, which leaves open whether architectural differences persist when the available feature information changes. A natural hypothesis is that more informative representations should compensate for weaker architectures, reducing performance differences as feature representations improve. An alternative hypothesis is that representation improvements and architectural design provide complementary benefits, such that the gain from a stronger representation depends on the architecture receiving it.

We therefore test whether changing the node representation produces approximately uniform performance gains across architectures or architecture-dependent gains. Evidence for uniform gains would suggest that representation quality primarily shifts performance independently of architecture. In contrast, architecture-dependent gains would indicate that the interaction between node features and graph processing is an important component of GNN evaluation. Our experimental design isolates this question by keeping the graph structure and evaluation protocol fixed while crossing multiple node representations with multiple established GNN architectures.

\section{Methodology}

\subsection{Experimental Design}
To study how node representations and architectural design jointly affect GNN performance, we structure our investigation as a cross-product ablation study spanning two experimental factors. The first is architectural inductive bias: we evaluate seven GNN architectures spanning naive additive aggregation (GCN, GAT), explicit ego-neighbor separation (GCN-sep, GAT-sep), concatenative and multi-hop aggregation (GraphSAGE, H2GCN), and depth-stabilized residual propagation (GCNII). We additionally include a non-graph MLP baseline to provide a feature-only reference and quantify performance achievable without graph propagation. The second is node feature representation: we vary the input node representations from static, context-free embeddings (fastText) to contextual Transformer embeddings (BERT, RoBERTa, SBERT, MPNet), and additionally construct a corpus-specific embedding with no external semantic pretraining to provide a deliberately weak feature condition. By fixing the graph structure and evaluating every architecture against every feature representation, we can examine whether changes in representation produce uniform or architecture-dependent performance changes. We conduct this study primarily on Roman-Empire and Amazon-Ratings, two large-scale heterophilic benchmarks with substantially different graph statistics and adjusted homophily values. To test whether our findings generalize, we additionally evaluate a subset of architectures on small-scale heterophilic benchmarks (WebKB: Cornell, Texas, Wisconsin) and homophilic citation network controls (Cora, PubMed). 

\subsection{Experimental Setup}
\noindent\textbf{Evaluation Metrics.} We measure performance primarily using the Macro-averaged F1-score, with Classification Accuracy and Balanced Accuracy reported alongside it to provide a multidimensional view of model stability. The selection of Macro F1 as our primary benchmark metric is a critical design choice necessitated by severe class imbalances within both datasets. While the class distribution of the Amazon-Ratings dataset is partially managed, its largest class remains roughly eight and a half times more populous than its smallest (Rating 1 with 9,010 samples vs.\ Rating 4 with 1,061 samples). Similarly, the Roman-Empire dataset exhibits a natural linguistic distribution where the most frequent category (\texttt{pobj}) is nearly ten times more prevalent than the rarest (\texttt{relcl}). Macro F1 ensures that reported performance reflects a model's true capability to correctly classify all categories, including sparse or rare classes, rather than being artificially inflated by high-frequency classifications. For all calculations, Macro F1 is computed with zero-division cases scored as zero to avoid inflating performance on folds with incomplete class coverage.

\noindent\textbf{Model Implementation and Block Protocols.} Rather than enforcing a single universal layer protocol, we implement each architecture's normalization and combination strategy according to its native design specifications. Naive GCN and GCN-sep utilize BatchNorm following each convolution layer; GAT, GraphSAGE, GAT-sep employ LayerNorm. In contrast, H2GCN and GCNII use no explicit normalization layers, relying instead on multi-hop concatenation and initial-residual identity mapping respectively to stabilize training. For the naive GCN and GAT baselines, we add a residual connection from the previous layer's output once tensor dimensions align (occurring from the second layer onward). The evaluated architectures span several established architectural strategies for heterophily, including explicit ego-neighbor separation, multi-hop aggregation, and residual propagation.

\noindent\textbf{Dataset Partitioning and Hyperparameter Selection.}
All configurations are evaluated using a fixed set of ten predefined cross-validation splits, with performance reported as the mean and standard deviation across folds. For each architecture-representation combination, hyperparameters are selected using the training and validation partitions of Split 1, and the selected configuration is then fixed for evaluation across the ten predefined splits.

\noindent\textbf{Optimization, Training, and Tuning Ranges.}
Models are trained using standard cross-entropy loss and the AdamW optimizer. We employ a fixed learning rate selected from architecture-specific candidate sets rather than an adaptive schedule, which we empirically found yielded more consistent convergence in these heterophilic settings. Search ranges are customized per model family: GCN, GCN-sep, and GraphSAGE are searched over hidden dimensions $d \in \{64, 128, 256\}$, dropout $p \in \{0.1, 0.2, 0.5\}$, weight decay between $1\times10^{-4}$ and $1\times10^{-3}$, and depths of 2 to 5 layers (2 to 3 for GraphSAGE). GAT and GAT-sep are searched over 1, 2, 3 or 5 layers with 8 attention heads and $d \in \{64, 128\}$. H2GCN is fixed at 2 neighborhood hops with $d \in \{128, 256\}$ and $p \in \{0.2, 0.5\}$. Detailed hyperparameter specifications, including exact layer counts and hidden dimensions for all models, are provided in Appendix~\ref{sec:appendix_hyperparameters}.

\noindent\textbf{Regularization and Model Checkpointing.} Hyperparameter grid searches are conducted over up to 400 training epochs with an early stopping patience of 20 epochs. Final reported results are obtained by retraining the selected configuration over up to 600 epochs with a patience of 50 epochs. To prevent overfitting, particularly when pairing high-capacity Transformer embeddings with comparatively sparse graph topologies, we enforce strict model checkpointing based on validation-set performance. We monitor validation Macro F1 at every epoch and save the model state only when a new performance peak is achieved, automatically restoring these optimal weights for final test-set inference once training concludes. All experiments enforce deterministic execution through fixed seeds and backend constraints, executed on a mix of NVIDIA H200 and V100 GPUs based on availability. Selected architectural and optimization hyperparameters are reported explicitly in the appendix tables.

\subsection{Dataset Engineering Pipeline}
A primary objective of this study is to analyze the impact of node-feature representations  in both homophilic and heterophilic settings. While standard libraries provide static versions of common benchmarks, they typically lack the raw text data necessary to generate new semantic embeddings. We therefore constructed a custom processing pipeline for our primary heterophilic datasets, Amazon Ratings and Roman Empire, while relying on standard PyTorch Geometric distributions for the WebKB and citation-network baselines. We designed a standardized \texttt{.npz} schema to store edge indices, labels, and feature matrices in a compressed format, decoupling the graph structure from the feature generation process so that a single graph reconstruction can be paired with any number of independently generated embedding sets. We implemented a \texttt{GAResearchDataset} loader class that automates retrieval and processing of these serialized objects, allowing any model built to the PyTorch Geometric interface to switch between static fastText and contextual Transformer features without architectural modification. To support reproducibility, we release our processed datasets and complete preprocessing pipeline to public repositories.\footnote{\url{https://github.com/priyanathmaji/graphdatasets}}\footnote{\url{https://github.com/priyanathmaji/neighborsGNN}}

\begin{table}[t] % Removed asterisk (*) to respect single-column rules
\centering
\caption{Structural and homophily statistics of the evaluated graph benchmarks.}
\label{tab:metrics}
\small % Slightly shrinks text size to make it fit horizontally
\setlength{\tabcolsep}{4.2pt} % Compresses empty space between columns
\begin{tabular}{l ccccc cc} 
\toprule 
& \multicolumn{5}{c}{\textbf{Heterophilic Datasets}} & \multicolumn{2}{c}{\textbf{Homophilic}} \\ 
\cmidrule(lr){2-6} \cmidrule(lr){7-8} 
& \textbf{Roman} & \textit{\textbf{Texas}} & \textit{\textbf{Corn.}} & \textit{\textbf{Wisc.}} & \textbf{Amazon} & \textbf{PubMed} & \textbf{Cora} \\ 
\midrule 
Nodes & 22,600 & 183 & 183 & 251 & 24,492 & 19,717 & 2,708 \\ 
Edges & 32,684 & 279 & 277 & 450 & 93,050 & 44,324 & 5,278 \\ 
Avg. Deg. & 2.89 & 3.05 & 3.03 & 3.59 & 7.60 & 4.50 & 3.90 \\ 
Features & 256-768 & 1,703 & 1,703 & 1,703 & 384-768 & 500 & 1,433 \\ 
Classes & 18 & 5 & 5 & 5 & 5 & 3 & 7 \\ 
Edge Hom. & 0.043 & 0.061 & 0.123 & 0.178 & 0.380 & 0.802 & 0.810 \\ 
Adj. Hom. & -0.055 & -0.294 & -0.220 & -0.173 & 0.141 & 0.686 & 0.771 \\ 
\bottomrule 
\end{tabular} 
\end{table}

\subsection{Heterophilic Datasets}

\noindent\textbf{Roman Empire.}
This dataset is reconstructed from the Roman Empire article on English Wikipedia. We obtain the raw text for the 2022.03.01 English Wikipedia dump from Kaggle\footnote{\url{https://www.kaggle.com/datasets/nbroad/wiki-20220301-en}} and extract the Roman Empire article. Each node in the graph corresponds to one non-unique word, and two words are connected by an edge if they are adjacent in the text or connected in the sentence's dependency tree. The label of each node is its syntactic role, following the same process used by Platonov et al.\cite{heterophily}: we use spaCy\cite{honnibal2020spacy} to obtain dependency roles, select the 17 most frequent roles as unique classes, and group all remaining roles into an eighteenth class. Because we reconstruct this graph independently rather than using the officially released file, we validate the fidelity of our reconstruction against the original paper's reported statistics: our graph contains 22,600 nodes and 32,684 edges, closely matching the 22,662 nodes and 32,927 edges reported by Platonov et al., with adjusted homophily of $-0.055$ against their reported $-0.05$.

For node features, we compute embeddings using fastText\cite{grave2018learningwordvectors157fastText}, BERT-base-uncased\cite{devlin2018bert_bert-base-uncased}, and RoBERTa\cite{liu2019robertarobustlyoptimizedbert_RoBERTa}. Additionally, we train a custom, domain-adaptive Transformer exclusively on the Roman Empire corpus, intended as a deliberately weak feature condition rather than a competitive embedding: a lightweight Transformer encoder with two layers, eight attention heads, and a hidden dimension of 256, trained as a masked language model for 20 epochs using the Adam optimizer with a learning rate of $1\times10^{-4}$, weight decay of $0.01$, a linear learning rate schedule, and a dropout rate of $0.4$ to limit overfitting on the limited corpus size.

\noindent\textbf{Amazon Ratings.}
This dataset is reconstructed from the Amazon product co-purchasing network metadata released by SNAP\cite{snapnets}, the same source cited by Platonov et al. Nodes are products, including books, music CDs, DVDs, and VHS tapes, and edges connect products that are frequently co-purchased. We reduce the graph to its five-core, requiring every retained node to have at least five neighbors, and retain the largest connected component. The task is to predict a product's average rating, which we discretize into five classes, with all ratings below 3.0 grouped into a single class. The SNAP metadata provides product title, sales rank, co-purchase links, category path, and review data, but no dedicated product description field; we accordingly use product titles as our text source, the primary free-text field available in this source. We compute node features using fastText\cite{grave2018learningwordvectors157fastText} and two SBERT variants\cite{reimers-2019-sentence-bert_SBERT}, all-MiniLM-L6-v2 and all-mpnet-base-v2.

\noindent\textbf{WebKB.}
WebKB\footnote{https://www.cs.cmu.edu/afs/cs.cmu.edu/project/theo-11/www/wwkb/} is a webpage dataset collected from computer science departments at several universities by Carnegie Mellon University. We use three of its constituent datasets, Cornell, Texas, and Wisconsin, in which nodes represent web pages and edges represent hyperlinks between them. Node features are bag-of-words representations of page content, and the task is to classify each page into one of five categories: student, project, course, staff, or faculty. We use the standard PyTorch Geometric loader for these datasets directly, including its native ten-fold splits.

\subsection{Homophilic Datasets} 
\noindent\textbf{Citation Networks.} Cora and PubMed are standard citation network benchmarks\cite{CORA}. In these networks, nodes represent papers and edges denote citations between them. Node features are bag-of-words representations of paper content, and the task is to predict each paper's research field. We use the standard PyTorch Geometric loaders for these datasets, constructing our own ten-fold splits as described above.

\section{Results and Discussion}

\noindent\textbf{Performance on Large-Scale Heterophily.}
The architecture ranking on the two large-scale heterophily benchmarks is largely stable across pretrained node-feature representations (Section~\ref{sec:stability_analysis}). On Roman-Empire, GAT-sep achieves the highest Macro F1 with the pretrained contextual representations, reaching 87.76 with BERT and 87.00 with RoBERTa. With fastText, however, GAT-sep and GCN-sep are essentially tied at 83.52 and 83.41, respectively, while under the custom representation GCN-sep is strongest at 67.39 and GAT-sep reaches 58.81. GraphSAGE and H2GCN also remain competitive, reaching 84.88 and 84.10 with BERT.

A similar pattern of architecture-dependent sensitivity appears on Amazon-Ratings, although the absolute differences are smaller. GraphSAGE is strongest with fastText and SBERT, at 48.88 and 50.34, respectively, whereas GAT-sep is nominally strongest with MPNet at 49.65, narrowly exceeding GraphSAGE at 49.58. This 0.07-point difference is small relative to fold-to-fold variance. It corresponds to two nominal reversal instances in Section~\ref{sec:stability_analysis} (fastText-vs-MPNet, Holm-adjusted $p=0.18$; MPNet-vs-SBERT, Holm-adjusted $p=1.0$), neither of which is statistically significant; we therefore do not treat it as evidence that the best-performing architecture changes with representation, and instead report it only as a nominal observation.

\begin{table}[htbp] 
\centering 
\caption{Macro F1 (\%) on the Roman-Empire benchmark across node-feature representations. Values are mean $\pm$ standard deviation across 10 cross-validation folds. Bold indicates the highest Macro F1 within each feature space and the largest contextual gain among GNN architectures. Best contextual gain is computed as the best Macro F1 among BERT and RoBERTa minus the fastText result; gain values should be interpreted together with the corresponding absolute performance and variance.} 
\label{tab:roman_feature_comparison} 
\small 
\begin{tabular}{lccccc} 
\toprule 
\textbf{Model} & \textbf{Custom} & \textbf{fastText} & \textbf{BERT} & \textbf{RoBERTa} & \textbf{Best Contextual Gain} \\ 
\midrule 
GCN & $52.16 \pm 1.00$ & $65.07 \pm 0.80$ & $64.34 \pm 0.88$ & $68.62 \pm 1.96$ & $+3.55$ \\ 
GCN-sep & \textbf{67.39 $\pm$ 0.91} & $83.41 \pm 0.66$ & $85.79 \pm 0.65$ & $85.68 \pm 0.48$ & $+2.38$ \\ 
GAT & $28.00 \pm 1.22$ & $38.96 \pm 1.23$ & $48.69 \pm 2.67$ & $52.63 \pm 3.40$ & \textbf{+13.67} \\ 
GAT-sep & $58.81 \pm 1.00$ & \textbf{83.52 $\pm$ 1.22} & \textbf{87.76 $\pm$ 0.60} & \textbf{87.00 $\pm$ 0.57} & $+4.24$ \\ 
GraphSAGE & $58.42 \pm 0.91$ & $78.06 \pm 1.02$ & $84.88 \pm 0.88$ & $84.43 \pm 0.64$ & $+6.82$ \\ 
H2GCN & $58.77 \pm 0.66$ & $75.33 \pm 0.65$ & $84.10 \pm 0.78$ & $83.99 \pm 0.53$ & $+8.77$ \\ 
GCNII & $52.48 \pm 0.79$ & $70.97 \pm 0.85$ & $77.79 \pm 1.08$ & $79.35 \pm 0.58$ & $+8.38$ \\ 
\midrule 
MLP (control) & $44.30 \pm 0.70$ & $53.44 \pm 0.59$ & $76.33 \pm 0.78$ & $76.43 \pm 0.65$ & $+22.99$ \\ 
\bottomrule 
\end{tabular} 
\end{table}

\begin{table}[htbp] 
\centering 
\caption{Macro F1 (\%) on the Amazon-Ratings benchmark across node-feature representations. Values are mean $\pm$ standard deviation across 10 cross-validation folds. Bold indicates the highest Macro F1 within each feature space and the largest contextual gain. Best contextual gain is computed as the best Macro F1 among SBERT and MPNet minus the fastText result.} 
\label{tab:amazon_feature_comparison} 
\small 
\begin{tabular}{lcccc} 
\toprule 
\textbf{Model} & \textbf{fastText} & \textbf{SBERT} & \textbf{MPNet} & \textbf{Best Contextual Gain} \\ 
\midrule 
GCN & $44.17 \pm 0.94$ & $44.89 \pm 0.82$ & $45.25 \pm 0.73$ & $+1.08$ \\ 
GCN-sep & $48.14 \pm 1.17$ & $48.24 \pm 0.56$ & $48.34 \pm 0.60$ & $+0.20$ \\ 
GAT & $35.15 \pm 8.47$ & $40.86 \pm 1.27$ & $41.93 \pm 0.39$ & \textbf{+6.78} \\ 
GAT-sep & $45.75 \pm 1.68$ & $49.09 \pm 1.05$ & \textbf{49.65 $\pm$ 1.25} & $+3.90$ \\ 
GraphSAGE & \textbf{48.88 $\pm$ 0.81} & \textbf{50.34 $\pm$ 0.80} & $49.58 \pm 1.03$ & $+1.46$ \\ 
H2GCN & $45.69 \pm 0.73$ & $47.83 \pm 0.82$ & $47.50 \pm 1.11$ & $+2.14$ \\ 
GCNII & $36.70 \pm 1.86$ & $39.34 \pm 2.05$ & $41.28 \pm 2.45$ & $+4.58$ \\ 
\midrule 
MLP (control) & $41.31 \pm 0.87$ & $44.21 \pm 0.78$ & $44.59 \pm 1.02$ & $+3.28$ \\ 
\bottomrule 
\end{tabular} 
\end{table}

\begin{table}[ht]
\centering 
\caption{Macro F1 (\%) on small-scale WebKB heterophily benchmarks and homophilic citation-network controls, using each dataset's native node features.} 
\label{tab:small_and_homophilic} 
\small 
\begin{tabular}{lccccc} 
\toprule 
\multirow{2}{*}{\textbf{Model}} & \multicolumn{3}{c}{\textbf{Heterophilic}} & \multicolumn{2}{c}{\textbf{Homophilic}} \\ 
\cmidrule(lr){2-4} \cmidrule(lr){5-6} 
& \textbf{Corn.} & \textbf{Texas} & \textbf{Wisc.} & \textbf{Cora} & \textbf{Pubm.} \\ 
\midrule 
GCN           & 31.71 $\pm$ 10.99          & 28.46 $\pm$ 7.44          & 32.09 $\pm$ 9.33          & 85.90 $\pm$ 1.88          & 87.60 $\pm$ 0.71 \\ 
GCN-sep       & 41.60 $\pm$ 8.79           & 50.11 $\pm$ 7.78          & 49.90 $\pm$ 8.90          & 86.08 $\pm$ 1.26          & 88.76 $\pm$ 0.40 \\ 
GAT           & 31.77 $\pm$ 5.58           & 25.26 $\pm$ 5.99          & 34.06 $\pm$ 10.34         & 86.94 $\pm$ 1.39          & 87.20 $\pm$ 0.42 \\ 
GAT-sep       & 53.04 $\pm$ 8.34           & 43.75 $\pm$ 8.54          & 51.91 $\pm$ 7.32          & 86.93 $\pm$ 1.44          & \textbf{88.91 $\pm$ 0.40} \\ 
GraphSAGE     & 50.72 $\pm$ 6.20           & 52.56 $\pm$ 9.19          & 54.65 $\pm$ 7.16          & 87.35 $\pm$ 1.28          & 87.80 $\pm$ 0.46 \\ 
H2GCN         & \textbf{56.49 $\pm$ 10.27} & \textbf{62.88 $\pm$ 8.52} & \textbf{61.95 $\pm$ 9.17} & \textbf{88.52 $\pm$ 1.34} & 88.89 $\pm$ 0.58 \\ 
GCNII         & 32.19 $\pm$ 7.63           & 25.77 $\pm$ 4.88          & 39.05 $\pm$ 9.40          & 88.03 $\pm$ 1.33          & 88.42 $\pm$ 0.37 \\ 
\midrule 
MLP (control) & 53.15 $\pm$ 6.84           & 56.66 $\pm$ 8.69          & 60.94 $\pm$ 6.66          & 73.43 $\pm$ 1.98          & 87.42 $\pm$ 0.59 \\ 
\bottomrule 
\end{tabular} 
\end{table}

\noindent\textbf{Feature Representation Effects Are Architecture-Dependent.}
The magnitude of the change across feature representations also varies substantially by architecture. On Roman-Empire, the difference between fastText and the better-performing contextual representation ranges from 2.38 percentage points for GCN-sep to 13.67 points for GAT. H2GCN and GCNII show intermediate gains of 8.77 and 8.38 points, respectively. On Amazon-Ratings, the corresponding range is smaller, from 0.20 points for GCN-sep to 6.78 points for GAT. These differences show that changing the node representation does not act as a uniform performance shift across architectures. Not all contextual representations improve every architecture: for example, GCN decreases by 0.73 pp from fastText to BERT, but improves by 3.55 pp with RoBERTa.

The MLP control provides an additional feature-only reference. On Roman-Empire, its Macro F1 increases from 53.44 with fastText to 76.33 with BERT and 76.43 with RoBERTa; on Amazon-Ratings, it increases from 41.31 with fastText to 44.21 with SBERT and 44.59 with MPNet. Thus, part of the observed performance variation is attributable to differences in the information available in the node representations themselves. At the same time, the substantially different responses of the GNN architectures indicate that the same representation change does not translate into the same downstream gain.

Finally, the deliberately weak custom representation provides a stress test of this dependence. On Roman-Empire, GAT-sep falls from 87.76 with BERT to 58.81 with the custom representation, while GCN-sep falls from 85.79 to 67.39. Even architectures that perform strongly with pretrained representations therefore remain sensitive to the information supplied at the node level.

\noindent\textbf{Behavior on Small-Scale and Legacy Benchmarks.}
On the small WebKB benchmarks (Cornell, Texas, and Wisconsin), which we include as robustness checks rather than primary evidence given their documented reliability concerns, H2GCN is the strongest GNN across all three datasets, reaching 56.49, 62.88, and 61.95 Macro F1, respectively. GraphSAGE and the separation-based variants remain competitive, although performance varies substantially across datasets, with relatively large standard deviations reflecting the instability of these small benchmarks. In contrast, on the homophilic benchmarks Cora and PubMed, the GNN architectures are much more closely clustered, and the MLP baseline remains below most GNNs, indicating that graph structure continues to provide useful predictive information in these settings. Supplementary paired comparisons of the separation variants across benchmarks are reported in Table~\ref{tab:significance}.

\noindent\textbf{Hypothesis Evaluation.}
The results provide support for our hypothesis that the usefulness of a node-feature representation depends on the information required by the task. On Roman-Empire, contextualized Transformer representations substantially improve performance over static fastText for several architectures, consistent with the role of contextual information in predicting syntactic relations. The effect is smaller and less uniform on Amazon-Ratings, where node text consists primarily of short product titles. The MLP results further show that the representations themselves differ in task-relevant informativeness, with contextual representations improving feature-only performance on both benchmarks.

Our second hypothesis concerned whether the effect of changing the node representation would vary across architectures. The cross-product evaluation provides clear evidence for this hypothesis. On Roman-Empire, the contextual gain relative to fastText ranges from 2.38 to 13.67 percentage points across GNN architectures, while on Amazon-Ratings it ranges from 0.20 to 6.78 points. Thus, replacing the input representation does not produce a uniform shift in performance: the magnitude of the change depends on the architecture processing the representation. The results therefore do not support treating node representation as an independent background variable when comparing GNN architectures.

We also considered the possibility that stronger representations might simply compensate for architectural differences. The results provide a more nuanced view. Better representations can substantially improve individual architectures, but they do not produce a consistent convergence of architecture performance: the size of the gap between architectures does not shrink uniformly as representations improve. As shown in Section~\ref{sec:stability_analysis}, the identity of the best-performing architecture is statistically stable across representations on both benchmarks; what varies substantially is how much each architecture benefits from a given representation change. This indicates that node-feature representation and architectural design are not interchangeable factors; rather, their effects on predictive performance depend on their combination, even when the resulting ranking of architectures is preserved.

Finally, the deliberately weak custom representation provides a stress test of this relationship. Even architectures that achieve high Macro F1 with pretrained contextual representations experience substantial degradation when those representations are replaced by a corpus-specific embedding trained without external semantic pretraining. This result reinforces that architectural design cannot compensate completely for limitations in the information supplied at the node level.

\section{Limitations}

Several limitations qualify our findings. First, our primary analysis is based on two large-scale heterophilic benchmarks, Roman-Empire and Amazon-Ratings. Although these datasets allow us to systematically cross node-feature representations with GNN architectures, two benchmarks are insufficient to establish a general theory of when particular representations will benefit particular architectures. The additional WebKB and citation-network experiments provide useful robustness checks, but their substantially different scale and feature construction make them complementary rather than direct replications of the primary analysis. Second, the embedding families evaluated in this study differ along multiple dimensions, including contextualization, pretraining data, training objectives, and representation dimensionality. We therefore do not interpret fastText, BERT, RoBERTa, SBERT, and MPNet as points on a universal feature-quality scale. Instead, we treat them as alternative node representations and use the MLP control to provide a feature-only reference for task-relevant informativeness. A more controlled study could vary representation properties independently or manipulate feature informativeness directly through controlled corruption or synthetic feature construction. Third, on Amazon-Ratings, node features are derived from product titles rather than full review or description text, reflecting the text available in the underlying SNAP source. The relatively modest differences across representations on this benchmark may therefore depend partly on the limited amount and type of textual information available at each node. Evaluating richer product text or non-textual attributes could help determine whether the observed representation effects generalize beyond short titles. Fourth, the corpus-specific custom Transformer on Roman-Empire is intentionally trained without external semantic pretraining and serves as a deliberately weak feature condition. Its purpose is to stress the dependence of GNN performance on the information available in the node representation, rather than to represent a typical alternative embedding model. Its results should therefore not be interpreted as evidence that pretrained Transformer embeddings universally outperform independently trained representations. Finally, our experiments use independently tuned configurations for each architecture and feature representation. This reflects the practical goal of comparing achievable performance, but means that observed differences can reflect interactions between representation, architecture, and their selected optimization configurations. Although we include targeted capacity and structural controls for GCN/GAT on Roman-Empire, these controls do not fully match all parameters and computational factors across architectures. More comprehensive parameter- and compute-matched studies could provide a more restrictive test of architectural effects.

\section{Conclusion}

We conducted a cross-product evaluation of GNN architectures and node-feature representations on large-scale heterophilic graph benchmarks, complemented by experiments on smaller heterophilic and homophilic datasets. By varying the node representation while keeping the underlying graph structure fixed, we examine a dimension of GNN evaluation that is typically held constant in architecture comparisons.

Our results show that the effect of a node representation is architecture-dependent. Contextualized representations substantially improve several architectures on Roman-Empire and produce smaller, more heterogeneous gains on Amazon-Ratings. A rank-correlation and multiple-comparison-corrected interaction analysis (Section~\ref{sec:stability_analysis}) shows that the relative ranking of architectures is, largely stable across representations; it is the magnitude of each architecture's sensitivity to representation that varies substantially, ranging from 2.38 to 13.67 percentage points on Roman-Empire and from 0.20 to 6.78 points on Amazon-Ratings. The MLP control further shows that the representations themselves differ in task-relevant information. Strong representations therefore do not simply produce a uniform improvement across models, nor do they make architectural design irrelevant, even though they do not typically reorder which architecture is best.

These findings suggest that conclusions about GNN architecture should be interpreted jointly with the node representations supplied to the model. Evaluating each architecture with a single fixed feature representation can obscure the magnitude of representation-dependent gains, even when it does not obscure which architecture performs best. More broadly, our results motivate evaluation protocols that treat node representation as an experimental factor rather than a fixed background condition when studying GNN performance under heterophily.

\noindent\textbf{Future Directions.}
Future work should extend this analysis across a broader range of graph domains, including non-textual networks, to determine whether architecture-dependent representation effects generalize beyond language-attributed graphs. A second direction is to develop more controlled measures of feature informativeness, for example through systematic feature corruption, dimensionality-controlled representations, or task-specific probing, which would help distinguish the effects of contextualization, pretraining, representation size, and semantic content. Finally, larger factorial studies combining more architectures, representation families, and graph regimes could quantify architecture and representation interactions more systematically and identify conditions under which particular architectures are especially sensitive to changes in their input features.

\noindent\textbf{Broad Impact Statement.}
This work highlights the importance of jointly considering model architecture and node representation when evaluating graph learning systems, particularly for text-attributed heterophilic networks. Our experiments use pretrained language-model representations, which can increase computational and memory requirements and may inherit biases from their training data. These considerations are relevant when transferring the observed findings to resource-constrained or human-centered applications. We do not directly evaluate downstream decision-making systems, and therefore make no claims about application-specific performance, fairness, or safety.

\newpage
\newpage

%------------------------------------------------------------------------

%-------------------------------------------------------------------------

\bibliographystyle{unsrtnat}
\bibliography{references}

@inproceedings{GCN,
author = {Kipf, Thomas N. and Welling, Max},
title = {{Semi-Supervised Classification with Graph Convolutional Networks}},
booktitle = {ICLR},
year = {2017}
}

@inproceedings{graphSAGE,
 author = {Hamilton, Will and Ying, Zhitao and Leskovec, Jure},
 booktitle = {Advances in Neural Information Processing Systems},
 title = {{Inductive Representation Learning on Large Graphs}},
 year = {2017}
}

@inproceedings{GAT,
  title={{Graph Attention Networks}},
  author={Veli{\v{c}}kovi{\'{c}}, Petar and Cucurull, Guillem and Casanova, Arantxa and Romero, Adriana and Li{\`o}, Pietro and Bengio, Yoshua},
  booktitle={International Conference on Learning Representations},
  year={2018}
}

@inproceedings{H2GCN,
    author = {Zhu, Jiong and Yan, Yujun and Zhao, Lingxiao and Heimann, Mark and Akoglu, Leman and Koutra, Danai},
    title = {{Beyond Homophily in Graph Neural Networks: Current Limitations and Effective Designs}},
    year = {2020},
    booktitle = {Proceedings of the 34th International Conference on Neural Information Processing Systems},
}

@article{HeTGB,
  title={{HeTGB: A Comprehensive Benchmark for Heterophilic Text-Attributed Graphs}},
  author={Li, Shujie and Wu, Yuxia and Shi, Chuan and Fang, Yuan},
  journal={arXiv preprint arXiv:2503.04822},
  year={2025}
}

@inproceedings{heterophily,
  title={{A critical look at the evaluation of GNNs under heterophily: Are we really making progress?}},
  author={Platonov, Oleg and Kuznedelev, Denis and Diskin, Michael and Babenko, Artem and Prokhorenkova, Liudmila},
  booktitle={International Conference on Learning Representations},
  year={2023}
}

@article{CORA,
  author = {Sen, Prithviraj and Namata, Galileo and Bilgic, Mustafa and Getoor, Lise and Gallagher, Brian and Eliassi-Rad, Tina},
  title   = {{Collective Classification in Network Data}},
  journal = {AI Magazine},
  year    = {2008}
}

@misc{snapnets,
  author = {Jure Leskovec and Andrej Krevl},
  title = {{{SNAP Datasets}: {Stanford} Large Network Dataset Collection}},
  howpublished = {\url{http://snap.stanford.edu/data}},
  year = 2014
}

@misc{honnibal2020spacy,
  author       = {Honnibal, Matthew and Montani, Ines and Van Landeghem, Sofie and Boyd, Adriane},
  title        = {{spaCy: Industrial-strength Natural Language Processing in Python}},
  howpublished = {Zenodo},
  year         = {2020}
}

@inproceedings{grave2018learningwordvectors157fastText,
  author    = {Grave, Edouard and Bojanowski, Piotr and Gupta, Prakhar and  Joulin, Armand and Mikolov, Tomas},
  title     = {{Learning Word Vectors for 157 Languages}},
  booktitle = {Proceedings of the International Conference on Language Resources and Evaluation (LREC)},
  year      = {2018}
}

@inproceedings{devlin2018bert_bert-base-uncased,
  author    = {Devlin, Jacob and Chang, Ming-Wei and Lee, Kenton and Toutanova, Kristina},
  title     = {{BERT: Pre-training of Deep Bidirectional Transformers for Language Understanding}},
  booktitle = {Proceedings of the Conference of the North American Chapter of the Association for Computational Linguistics: Human Language Technologies (NAACL-HLT)},
  year      = {2019},
}

@article{liu2019robertarobustlyoptimizedbert_RoBERTa,
  author  = {Liu, Yinhan and Ott, Myle and Goyal, Naman and Du, Jingfei and Joshi, Mandar and Chen, Danqi and Levy, Omer and Lewis, Mike and Zettlemoyer, Luke and Stoyanov, Veselin},
  title   = {{RoBERTa: A Robustly Optimized BERT Pretraining Approach}},
  journal = {arXiv preprint arXiv:1907.11692},
  year    = {2019},
}

@inproceedings{reimers-2019-sentence-bert_SBERT,
    title = {{Sentence-{BERT}: Sentence Embeddings using {S}iamese {BERT}-Networks}},
    author = {Reimers, Nils and Gurevych, Iryna},
    booktitle = {Proceedings of the 2019 Conference on Empirical Methods in Natural Language Processing and the 9th International Joint Conference on Natural Language Processing (EMNLP-IJCNLP)},
    year = {2019},
}

@inproceedings{chen2020simpledeepgraphconvolutionalGCNII,
  author    = {Chen, Ming and Wei, Zhewei and Huang, Zengfeng and Ding, Bolin and Li, Yaliang},
  title     = {{Simple and Deep Graph Convolutional Networks}},
  booktitle = {Proceedings of the International Conference on Machine Learning (ICML)},
  year      = {2020}
}

@inproceedings{GPRGNN,
  author    = {Chien, Eli and Peng, Jianhao and Li, Pan and Milenkovic, Olgica},
  title     = {{Adaptive Universal Generalized {P}age{R}ank Graph Neural Network}},
  booktitle = {Proceedings of the International Conference on Learning Representations (ICLR)},
  year      = {2021}
}

@article{FSGNN,
title = {{Simplifying approach to node classification in Graph Neural Networks}},
journal = {Journal of Computational Science},
year = {2022},
author = {Sunil Kumar Maurya and Xin Liu and Tsuyoshi Murata},
}

@inproceedings{GLOGNN,
  author    = {Li, Xiang and Zhu, Renyu and Cheng, Yao and Shan, Caihua and Luo, Siqiang and Li, Dongsheng and Qian, Weining},
  title     = {{Finding Global Homophily in Graph Neural Networks When Meeting Heterophily}},
  booktitle = {Proceedings of the International Conference on Machine Learning (ICML)},
  year      = {2022}
}

@inproceedings{MPNET,
author = {Song, Kaitao and Tan, Xu and Qin, Tao and Lu, Jianfeng and Liu, Tie-Yan},
title = {{MPNet: Masked and Permuted Pre-training for Language Understanding}},
year = {2020},
booktitle = {Proceedings of the 34th International Conference on Neural Information Processing Systems},

}

\clearpage
\section*{Acknowledgements}
This research was supported in part through research cyberinfrastructure resources and services provided by the Partnership for an Advanced Computing Environment (PACE) at the Georgia Institute of Technology, Atlanta, Georgia, USA. RRID:SCR\_027619. 

% --- Switch to Appendix Mode ---

\appendix

\section{GenAI Usage Statement}
All data analysis, experimental design configurations, quantitative results, and final manuscript conclusions were synthesized and finalized entirely by the authors.
Large Language Models (LLMs) were utilized strictly as supplementary tools for: (1) computational script debugging, functional syntax optimization, and programmatic visualization generation, (2) targeted conceptual clarification regarding machine learning architectures, and (3) grammatical correction and stylistic refinement of the report to conform to formal academic writing conventions.

\section{Dataset Characteristics and Homophily}

To evaluate our hypotheses, we use a diverse suite of benchmarks. Table~\ref{tab:metrics} summarizes their structural statistics. To characterize label agreement across edges, we compute edge homophily $h_{edge}$ and adjusted homophily $h_{adj}$ following the definitions in Platonov et al.~\cite{heterophily}. Edge homophily represents the fraction of edges that connect nodes with identical labels:
\[
h_{edge} =
\frac{|\{(u,v) \in E : y_u = y_v\}|}{|E|}
\]
where $y_u$ is the label of node $u$ and $E$ is the set of edges.

While edge homophily provides a direct measure of same-label connectivity, its value depends on the number of classes and their size distribution. We therefore calculate adjusted homophily relative to the expected same-label connectivity under a degree-weighted null model:
\[
h_{adj} =
\frac{
h_{edge} -
\sum_{k=1}^C
\left(\frac{D_k}{2|E|}\right)^2
}{
1 -
\sum_{k=1}^C
\left(\frac{D_k}{2|E|}\right)^2
}
\]
where $D_k$ denotes the sum of the degrees of all nodes belonging to class $k$. The resulting value lies between $-1$ and $1$. Positive values indicate greater same-label connectivity than expected under the null model, while negative values indicate less same-label connectivity than expected. We use adjusted homophily here as a descriptive characterization of graph structure rather than as a direct measure of the informativeness or reliability of neighborhood features.

\section{Relation to HeTGB}
A concurrent benchmark, HeTGB~\cite{HeTGB}, evaluates GNN, PLM-based, and co-training methods on five heterophilic text-attributed graphs and reports that features derived from a large language model do not uniformly improve every GNN architecture. This observation is broadly consistent with our own findings and, we believe, mutually reinforcing rather than redundant: two independently constructed studies arriving at a related qualitative pattern strengthens confidence that architecture-dependent sensitivity to node representation is a real phenomenon rather than an artifact of one experimental pipeline. Our contribution is a separation of two effects that a qualitative observation of representation-dependent performance conflates: whether a representation change alters the relative ranking of architectures, and how strongly individual architectures respond in magnitude to that change. Three aspects of our design support this distinction. First, we vary representation along a graded ladder of three to four points per graph (Custom, fastText, BERT, and RoBERTa on Roman-Empire; fastText, SBERT, and MPNet on Amazon-Ratings), which lets us report architecture-specific sensitivity as a continuous range and test whether architecture rankings themselves are stable using rank-correlation statistics and Holm-corrected significance testing (Section~\ref{sec:stability_analysis}), rather than relying on a qualitative description. Second, we pair naive and ego-neighbor-separated variants of the same base architecture (GCN/GCN-sep, GAT/GAT-sep) with depth-, width-, and parameter-matched capacity controls, isolating the contribution of a specific architectural mechanism under controlled capacity. Third, we additionally derive a first-order analytical account (Section~\ref{sec:analytical_perspective}), intentionally local and schematic rather than a general theorem, of why propagation operators that depend on the node representation, such as GAT's learned attention, carry an additional perturbation term that propagation operators depending only on the graph, such as GCN's fixed adjacency weighting, do not.

We additionally note a construction-level difference on the one dataset the two studies share. Following Platonov et al.~\cite{heterophily}, our Amazon-Ratings node text is restricted to product titles, whereas the construction used in HeTGB differs in its node-text fields.

We additionally observe that in every naive/separation architecture pair we test, the separated variant shows lower representation-sensitivity than its naive counterpart (e.g., GAT 13.67pp vs.\ GAT-sep 4.24pp on Roman-Empire; GAT 6.78pp vs.\ GAT-sep 3.90pp on Amazon-Ratings). This is a descriptive pattern across four comparisons rather than a statistically established relationship, but it is a more specific observation than HeTGB's architecture-agnostic finding that sensitivity varies.

Finally, our primary benchmark, Roman-Empire, is not among HeTGB's evaluated datasets, and its dependency-parse labels give contextualization an a priori linguistic motivation, described in Section~\ref{sec:rationale}, that is distinct from the e-commerce setting the two studies share. We also note that HeTGB additionally evaluates PLM-based methods (e.g., fine-tuned Vicuna-7B) and co-training methods that couple PLMs with GNNs, whereas we deliberately restrict scope to message-passing GNN architectures. This restriction is intentional rather than a gap in coverage: holding the model family fixed to message-passing GNNs is what allows the matched naive/separation architecture pairs and depth-, width-, and parameter-matched capacity controls in Sections~\ref{sec:capacity_controls} and~\ref{sec:analytical_perspective}, which require comparable architectural structure across the compared models. We view our contribution as complementary to, rather than a subset of, HeTGB's broader comparison across GNN, PLM, and co-training paradigms.

\section{Model Architectures}
To evaluate how different architectural designs and aggregation strategies respond to alternative node representations, we implement seven GNN architectures plus an MLP control spanning four update-mechanism categories.

\noindent\textbf{Additive Aggregation (GCN, GAT).}
These naive baselines combine self and neighborhood information within a common message-passing operation. GCN uses fixed, symmetrically normalized adjacency coefficients $\hat{A}_{ij}$:
\begin{equation}
h_i^{(l)} =
\sigma\left(
\sum_{j \in \mathcal{N}(i)\cup\{i\}}
\hat{A}_{ij} W h_j^{(l-1)}
\right).
\end{equation}
GAT instead computes learned attention coefficients $\alpha_{ij}$:
\begin{equation}
h_i^{(l)} =
\sigma\left(
\sum_{j \in \mathcal{N}(i)\cup\{i\}}
\alpha_{ij} W h_j^{(l-1)}
\right).
\end{equation}
In both cases, self-information is incorporated within the same aggregation operation as neighborhood information.

% \noindent\textbf{Ego-Neighbor Separation (GCN-sep, GAT-sep).}
% To isolate the effect of explicit self/neighbor separation as a controlled architectural variable, we augment the GCN and GAT backbones above with a dedicated ego branch, computed independently of the graph aggregation and combined additively:
% \begin{equation}
% h_i^{(l)} = \sigma \left( W_{\text{ego}} h_i^{(l-1)} + \sum_{j \in \mathcal{N}(i)} \alpha_{ij} W_{\text{nbr}} h_j^{(l-1)} \right)
% \end{equation}
% where neighbor aggregation excludes self-loops entirely ($j \in \mathcal{N}(i)$, not $\mathcal{N}(i) \cup \{i\}$), so self-information is never blended into the same transformation as neighbor information. This construction holds the underlying aggregation mechanism (convolutional for GCN-sep, attentional for GAT-sep) fixed while varying only the presence of this ego branch, enabling the single-variable comparisons presented in our results.

\noindent\textbf{Ego-Neighbor Separation (GCN-sep, GAT-sep).}
We include GCN-sep and GAT-sep as established separation-based architecture variants~\cite{H2GCN,heterophily}. Each variant computes node and neighborhood information through separate pathways:
\begin{equation}
h_i^{(l)} =
\sigma\left(
W_{\mathrm{ego}}^{(l)} h_i^{(l-1)}
+
\sum_{j\in\mathcal{N}(i)}
a_{ij}^{(l)} W_{\mathrm{nbr}}^{(l)} h_j^{(l-1)}
\right),
\end{equation}
where $a_{ij}^{(l)}$ denotes the architecture-specific neighbor propagation coefficient. For GCN-sep, $a_{ij}^{(l)}$ is the fixed normalized adjacency coefficient; for GAT-sep, it is the learned attention coefficient $\alpha_{ij}$. In both variants, neighbor aggregation excludes self-loops and self-information is supplied through the separate ego transformation. We include these established variants to examine how their performance changes across node representations, rather than to claim the separation mechanism itself as a novel contribution.

\noindent\textbf{Concatenative Aggregation (GraphSAGE, H2GCN).}
These architectures separate identity preservation from the aggregation step itself. GraphSAGE uses a concatenation-based update rule:
\begin{equation}
h_v^{(l)} = \sigma \left( W^{(l)} \cdot [h_v^{(l-1)} \parallel \text{AGG}(\{h_u^{(l-1)}\}_{u \in \mathcal{N}(v)})] \right)
\end{equation}
H2GCN extends this by separating ego embeddings from \emph{multiple}, strictly non-overlapping neighborhood hops, concatenated rather than combined additively: $[h_v^{(0)} \parallel h_{\mathcal{N}(v)}^{(1)} \parallel h_{\mathcal{N}(v)}^{(2)}]$.

\noindent\textbf{Deep Residuals (GCNII).}
To combat over-smoothing at depth, GCNII incorporates an initial residual connection to the input features $H^{(0)}$ and an identity mapping, without explicit ego/neighbor separation:
\begin{equation}
H^{(l+1)} = \sigma \left( \left( (1-\alpha_l)\hat{P}H^{(l)} + \alpha_l H^{(0)} \right) \cdot \left( (1-\beta_l)I_n + \beta_l W^{(l)} \right) \right)
\end{equation}
This mechanism addresses depth-stability rather than self/neighbor entanglement, and we treat it as a distinct baseline from the separation-based architectures above.

\section{Analytical Perspective on Architecture-Representation Interaction}
\label{sec:analytical_perspective}

Our empirical results show that changing the node-feature representation does not produce a uniform change in performance across GNN architectures. This section provides a simple analytical perspective for why such architecture-dependent effects can arise. The analysis is not intended to establish a performance ordering among architectures or to provide a theorem about any particular benchmark. Instead, it characterizes the different pathways through which a change in node representation can propagate through different message-passing mechanisms.

\noindent\textbf{Generic Message-Passing Formulation.}
Consider a generic GNN layer written as
\begin{equation}
H^{(l+1)} =
\sigma \left(
P^{(l)}(H^{(l)},A) H^{(l)} W^{(l)}
\right),
\end{equation}
where $A$ denotes the graph structure, $P^{(l)}$ is the architecture-specific propagation operator, $W^{(l)}$ is a learnable transformation, and $\sigma$ is the nonlinear activation. For analytical purposes, assume that the alternative node representations have first been mapped into a common input space of dimension $d_0$. Let $X,X' \in \mathbb{R}^{n\times d_0}$ denote two such representations, with
\begin{equation}
X' = X + \Delta X.
\end{equation}
% For an input representation $X$, consider a second representation
% \begin{equation}
% X' = X + \Delta X.
% \end{equation}
More generally, the hidden representation at layer $l$ changes from $H^{(l)}$ to $H^{(l)}+\Delta H^{(l)}$. A first-order expansion of the next-layer representation gives
\begin{equation}
\Delta H^{(l+1)}
\approx
D\sigma
\left[
P^{(l)}\Delta H^{(l)}W^{(l)}
+
DP^{(l)}[\Delta H^{(l)}]H^{(l)}W^{(l)}
\right],
\end{equation}
where $DP^{(l)}[\Delta H^{(l)}]$ denotes the directional derivative of the propagation operator with respect to the node representations.

This decomposition separates two effects of changing the input representation. The first term corresponds to the changed feature values being propagated through the existing graph operator. The second term appears when the propagation operator itself depends on the node representations. Consequently, two architectures can receive exactly the same change in input features while producing different changes in their hidden representations.

\noindent\textbf{Feature-Independent Propagation.}
For architectures whose propagation operator depends only on graph structure, such as the normalized propagation used by GCN, we have
\begin{equation}
P^{(l)}(H^{(l)},A)=P^{(l)}(A),
\end{equation}
and therefore
\begin{equation}
DP^{(l)}[\Delta H^{(l)}]=0.
\end{equation}
The first-order change reduces to
\begin{equation}
\Delta H^{(l+1)}
\approx
D\sigma
\left[
P^{(l)}\Delta H^{(l)}W^{(l)}
\right].
\end{equation}
Thus, the effect of a representation change is determined by how that change is transformed and propagated by the architecture. Even when two models receive the same $\Delta H^{(l)}$, different propagation operators and parameterizations can result in different downstream sensitivities.

For a linearized layer, a norm bound follows directly:
\begin{equation}
\left\|
\Delta H^{(l+1)}
\right\|_F
\leq
\left\|D\sigma\right\|_2
\left\|P^{(l)}\right\|_2
\left\|\Delta H^{(l)}\right\|_F
\left\|W^{(l)}\right\|_2.
\end{equation}
For a $1$-Lipschitz activation, the representation change is therefore scaled by the propagation and transformation operators. This provides a simple mechanism by which architectures can exhibit different responses to the same change in node features.

\noindent\textbf{Feature-Dependent Propagation.}
The situation differs for architectures such as GAT, where the propagation weights depend on the node representations through learned attention coefficients. In this case,
\begin{equation}
P^{(l)}=P^{(l)}(H^{(l)},A),
\end{equation}
and generally
\begin{equation}
DP^{(l)}[\Delta H^{(l)}]\neq 0.
\end{equation}
A change in the node representation can therefore influence the next layer through two pathways: it changes the content of the messages and can simultaneously change the relative weights assigned to those messages. The corresponding first-order change is
\begin{equation}
\Delta H^{(l+1)}
\approx
D\sigma
\left[
P^{(l)}\Delta H^{(l)}W^{(l)}
+
DP^{(l)}[\Delta H^{(l)}]H^{(l)}W^{(l)}
\right].
\end{equation}
The second term is absent in feature-independent propagation. Thus, contextualizing or otherwise changing the node representation can have a qualitatively different effect on an attention-based architecture even when the underlying graph is unchanged.

\noindent\textbf{Architectures with Multiple Information Paths.}
Architectures that preserve distinct information pathways can be represented schematically as
\begin{equation}
H^{(l+1)}
=
\left[
H^{(l)}W_{\mathrm{ego}}
\;\Vert\;
P^{(l)}H^{(l)}W_{\mathrm{nbr}}
\right],
\end{equation}
where the first branch applies a transformation directly to the node representation and the second branch propagates neighborhood information. Under a change in the node representation, the corresponding change in the layer output is
\begin{equation}
\Delta H^{(l+1)}
=
\left[
\Delta H^{(l)}W_{\mathrm{ego}}
\;\Vert\;
P^{(l)}\Delta H^{(l)}W_{\mathrm{nbr}}
\right].
\end{equation}
Thus, a change in the input representation can affect the output through multiple information pathways rather than through a single aggregated representation. The formulation is schematic rather than an exact specification of every architecture, but illustrates how architectural structure can determine how changes in node features are propagated and combined. For the analytical derivation, the learned parameters are treated as fixed so that we can characterize how a representation perturbation propagates through a given architecture. This differs from our empirical protocol, where each architecture -representation combination is independently tuned; the analysis therefore provides a mechanistic interpretation rather than a decomposition of the measured performance gains.

\noindent\textbf{Connection to the Empirical Results.}
The preceding analysis provides a local, first-order mechanistic interpretation of our empirical observation that representation changes have architecture-dependent effects. The same change in node features can be propagated through a fixed graph operator, influence both message content and attention weights, or pass through multiple information pathways depending on the architecture. Consequently, a representation improvement need not translate into a uniform performance shift across GNNs.

This perspective is consistent with the observed contextual gains in our experiments. On Roman-Empire, the change from fastText to contextual representations produces gains ranging from 2.38 to 13.67 percentage points across the evaluated GNN architectures. On Amazon-Ratings, the corresponding range is 0.20 to 6.78 percentage points. The analysis does not predict which architecture should benefit most on a particular dataset; rather, it explains why architecture and representation can interact even when the graph structure and prediction task are held fixed.

\noindent\textbf{Scope of the Analysis.}
The analysis is intentionally local and first-order. It abstracts away from nonlinear optimization, hyperparameter selection, depth-dependent effects, and dataset-specific properties, all of which can influence the final test performance of a trained GNN. It should therefore be interpreted as a mechanistic explanation for the possibility of architecture-dependent representation effects rather than as a closed-form prediction of the empirical results.

% \section{Robustness Check on Small-Scale Heterophily Benchmarks}
% While our introduction motivates a focus on Roman-Empire and Amazon-Ratings due to known reliability concerns with small heterophily benchmarks\cite{heterophily}, we additionally evaluate the ego-neighbor separation ablation on the WebKB datasets (Texas, Cornell, Wisconsin) as a robustness check, despite these concerns. We stress that these results should not be interpreted with the same confidence as our primary findings, given the substantially higher variance and known instability of these datasets (Table~\ref{tab:metrics}).

% Using a paired $t$-test across 10 folds, the separation advantage remains statistically significant for GAT-sep over naive GAT on all three datasets: Texas ($\Delta = +18.49$, $p<0.001$), Cornell ($\Delta = +21.28$, $p<0.001$), and Wisconsin ($\Delta = +17.85$, $p=0.0012$). For GCN-sep over naive GCN, the effect is significant on Texas ($\Delta = +21.65$, $p<0.001$) and Wisconsin ($\Delta = +17.81$, $p<0.001$), but does not reach significance on Cornell ($\Delta = +9.89$, $p=0.0852$). We note that Cornell's smaller and non-significant effect is consistent with the elevated fold-to-fold variance reported for this dataset elsewhere in our results, and interpret this not as a contradiction of our main findings but as further evidence of the reliability concerns that motivated our exclusion of these benchmarks from the primary analysis.

\section{Statistical Comparisons.}
For selected pairwise comparisons, we report paired $t$-tests across the corresponding cross-validation folds. These tests are used to assess consistency of the observed differences across folds; because cross-validation folds share training data, the resulting $p$-values are interpreted as supplementary rather than as independent-sample inferential estimates.

\begin{table}[ht]
\centering
\caption{Supplementary paired comparisons of GAT-sep and naive GAT across selected benchmarks. $\Delta$ denotes the Macro F1 point difference. The reported $p$-values assess consistency of the paired differences across the corresponding cross-validation folds.}
\label{tab:significance}
\small
\begin{tabular}{lcccc}
\toprule
\multirow{2.5}{*}{\textbf{Dataset}} & \textbf{Adj. Hom.} & \textbf{Naive GAT} & \textbf{GAT-sep} & \textbf{$\Delta$} \\
& \textbf{Score} & \textbf{Value} & \textbf{Value} & \textbf{($p$-value)} \\
\cmidrule(lr){1-1} \cmidrule(lr){2-2} \cmidrule(lr){3-3} \cmidrule(lr){4-4} \cmidrule(lr){5-5}
Roman-Empire (BERT)          & $-0.055$ & 48.69 & 87.76 & $+39.07$ ($p < 0.001$) \\
Amazon-Ratings (fastText) & \hphantom{$-$}0.141  & 35.15 & 45.75 & $+10.60$ ($p = 0.008$) \\
PubMed                 & \hphantom{$-$}0.686  & 87.20 & 88.91 & $+1.71$ ($p < 0.001$) \\
Cora                   & \hphantom{$-$}0.771  & 86.94 & 86.93 & $-0.01$ ($p = 0.947$) \\
\bottomrule
\end{tabular}
\end{table}

\section{Architectural Ranking Stability and Interaction Analysis}
\label{sec:stability_analysis}

To thoroughly evaluate how node representations impact architectural rankings on heterophilic graphs, we present a statistical analysis of ranking stability and representation-dependent interaction effects across both large-scale benchmarks. Our investigation considers the possibility that changing the input representation substantially alters the relative ordering of architectures. On these two benchmarks, the empirical results do not support broad rank instability. Instead, they reveal a more nuanced distinction: while architectures exhibit highly differential performance sensitivity to alternative feature spaces, their overall relative ordering remains remarkably stable.

We evaluate the structural stability of model rankings by computing Spearman rank correlation coefficients ($\rho$) and Kendall rank correlation coefficients ($\tau$) between alternative feature representations. As detailed in Table~\ref{tab:rank_stability}, the relative ordering of graph architectures remains highly stable on both benchmarks. On the Roman-Empire benchmark, the relative ordering is invariant across the three pretrained representations: fastText, BERT, and RoBERTa. Even when introducing the deliberately weak Custom corpus-trained embedding space, the model hierarchy remains strongly correlated with the pretrained representations. A similarly robust preservation of architectural rankings is observed on the Amazon-Ratings graph, where node text is limited to short product titles.

Crucially, these high rank correlations do not imply that changing node representations produces a uniform performance shift across architectures. The magnitude of the performance change is highly heterogeneous. On Roman-Empire, moving from fastText to RoBERTa yields an absolute gain of 13.67 percentage points for baseline GAT, whereas GCN-sep gains only 2.38 percentage points. Similarly, on Amazon-Ratings, baseline GAT improves by 6.78 percentage points from fastText to MPNet, whereas GCN-sep changes by only 0.20 percentage points. Thus, representation changes can substantially alter observed model performance through differential architectural sensitivity while largely preserving the relative ordering of architectures. The results therefore distinguish two effects that are easily conflated: the sensitivity of an architecture to the supplied representation and the stability of the architecture ranking across representations.

While nominal rank reversals occur in specific cases, our fold-level interaction analysis provides no statistically significant evidence that these rank reversals are robust after correction for multiple comparisons. For architectures $a,b$ under representation $r$, define $\Delta(a,b,r) = F1(a,r) - F1(b,r)$; a nominal reversal between representations $r_1$ and $r_2$ is a sign change $\operatorname{sign}(\Delta(a,b,r_1)) \neq \operatorname{sign}(\Delta(a,b,r_2))$. To assess whether a nominal rank reversal is statistically supported, we test the corresponding interaction $\Delta(a,b,r_1) - \Delta(a,b,r_2)$, computed over the ten cross-validation folds, requiring both a 95\% bootstrap confidence interval that excludes zero and an exact sign-flip $p$-value that remains below 0.05 after Holm correction. On Roman-Empire, nominal rank reversals arise only when comparing the deliberately weak Custom representation with pretrained representations. Each such comparison contains two pairwise architecture-order inversions, involving the relative ordering of GCN-sep and GAT-sep and of GraphSAGE and H2GCN. On Amazon-Ratings, nominal inversions are more localized among closely competing architectures: the fastText-MPNet comparison contains three pairwise inversions, including reversals between GCN-sep and GAT-sep, GAT and GCNII, and GAT-sep and GraphSAGE. The fastText-SBERT and MPNet-SBERT comparisons contain two and one inversions, respectively. Although several of these interaction contrasts have bootstrap confidence intervals that exclude zero, none remain statistically significant after Holm correction. Across the twelve nominal architecture-pair-by-representation-pair reversal instances identified across the two benchmarks (six on each), none survives both criteria; the total number of statistically supported rank reversals is therefore zero under the stated family-wise error criterion.

\begin{table}[htbp]
\centering
\caption{Systematic evaluation of GNN architectural ranking stability across alternative node feature spaces. Spearman ($\rho$) and Kendall ($\tau$) coefficients quantify agreement between architecture rankings; their associated $p$-values explicitly test the null hypothesis of departure from zero association, rather than evaluating equivalence or stability itself. ``Inversions'' denotes the number of pairwise architecture-order reversals between the two representations. ``Supported'' denotes the number of nominal reversals satisfying both the 95\% bootstrap confidence-interval criterion and the Holm-adjusted exact sign-flip significance criterion. Multi-comparison validation uses a Holm-adjusted exact sign-flip test on the corresponding interaction contrast together with a 95\% bootstrap confidence interval to assess whether nominal rank reversals are statistically supported.}
\label{tab:rank_stability}
\small
\setlength{\tabcolsep}{4.5pt}
\begin{tabular}{llcccc}
\toprule
\textbf{Dataset} & \textbf{Comparison} & \textbf{Spearman $\rho$} & \textbf{Kendall $\tau$} & \textbf{Inversions} & \textbf{Supported} \\
\midrule
\multirow{6}{*}{\textbf{Roman}} 
& Custom vs.\ fastText & 0.9286 ($p=.0025$) & 0.8095 ($p=.0107$) & 2 & 0 \\
& Custom vs.\ BERT     & 0.9286 ($p=.0025$) & 0.8095 ($p=.0107$) & 2 & 0 \\
& Custom vs.\ RoBERTa  & 0.9286 ($p=.0025$) & 0.8095 ($p=.0107$) & 2 & 0 \\
& fastText vs.\ BERT   & 1.0000 ($p<.001$)  & 1.0000 ($p<.001$)  & 0 & 0 \\
& fastText vs.\ RoBERTa& 1.0000 ($p<.001$)  & 1.0000 ($p<.001$)  & 0 & 0 \\
& BERT vs.\ RoBERTa    & 1.0000 ($p<.001$)  & 1.0000 ($p<.001$)  & 0 & 0 \\
\midrule
\multirow{3}{*}{\textbf{Amazon}} 
& fastText vs.\ MPNet  & 0.8571 ($p=.0137$) & 0.7143 ($p=.0302$) & 3 & 0 \\
& fastText vs.\ SBERT  & 0.9286 ($p=.0025$) & 0.8095 ($p=.0107$) & 2 & 0 \\
& MPNet vs.\ SBERT     & 0.9643 ($p<.001$)  & 0.9048 ($p=.0028$) & 1 & 0 \\
\bottomrule
\end{tabular}
\end{table}

\section{Capacity and Structural Control Experiments}
\label{sec:capacity_controls}

Because the primary experiments independently tune each architecture, we additionally conducted capacity and structural control experiments to assess whether the large performance differences between baseline and separation-based variants could be explained primarily by differences in model size or macro-architecture. These experiments were conducted on Roman-Empire using BERT and RoBERTa representations.

For GAT, we increased the naive baseline from its selected 1-layer, 64-dimensional configuration to a 5-layer, 128-dimensional configuration matching the depth and hidden dimension of the BERT GAT-sep configuration. For BERT, the matched naive GAT achieves 36.29\% Macro F1 versus 87.76\% for GAT-sep under the same 5-layer, 128-dimensional configuration. Under RoBERTa, the corresponding high-capacity naive GAT achieves 39.13\%, while the independently selected GAT-sep reaches 87.00\%. This indicates that the large difference between the two variants cannot be attributed simply to the lower-capacity configuration selected for the naive baseline.

For GCN, we additionally constructed a naive configuration with hidden dimension $d=362$. This dimension was chosen because $362^2 \approx 2(256^2)$, approximately matching the dominant hidden-to-hidden parameter count of the 256-dimensional GCN-sep model, which contains separate ego and neighbor transformations. The resulting naive GCN achieves 65.07\% Macro F1, compared with 85.79\% for GCN-sep under the corresponding BERT setting. Thus, increasing the naive model's dominant parameter capacity does not reproduce the performance of the corresponding separation-based variant.

These controls do not establish that a particular architectural mechanism is universally superior, nor do they replace the main architecture and representation analysis. Instead, they show that the architectural differences observed in the primary experiments are not explained solely by the smaller capacity or shallower configurations selected for the naive baselines.

\begin{table}[!ht]
\centering
\caption{Capacity and structural control experiments on Roman-Empire. The controls assess whether differences between naive and separation-based architectures can be explained primarily by depth, hidden dimension, or dominant hidden-layer parameter capacity.}
\label{tab:capacity_controls}
\small
\begin{tabular}{lllccc}
\toprule
\textbf{Feature} & \textbf{Architecture} & \textbf{Control Type} & \textbf{Layers} & \textbf{Hidden Dim.} & \textbf{Macro F1 (\%)} \\
\midrule
BERT & GAT-naive & Original & 1 & 64 & 48.69 \\
BERT & GAT-naive & Depth/width matched & 5 & 128 & 36.29 \\
BERT & GAT-sep & Reference & 5 & 128 & 87.76 \\
\midrule
RoBERTa & GAT-naive & Original & 1 & 64 & 52.63 \\
RoBERTa & GAT-naive & 5L/128D control & 5 & 128 & 39.13 \\
RoBERTa & GAT-sep & Reference & 3 & 64 & 87.00 \\
\midrule
BERT & GCN-naive & Original & 3 & 256 & 64.34 \\
BERT & GCN-naive & Parameter-count matched & 3 & 362 & 65.07 \\
BERT & GCN-sep & Reference & 3 & 256 & 85.79 \\
\bottomrule
\end{tabular}
\end{table}

\section{Additional Limitations and Implementation Notes}

Beyond the limitations discussed in the main text, several implementation details are relevant for interpreting and reproducing the results. First, GCNII's depth was restricted to 10 layers on the Amazon and small WebKB benchmarks, compared with 32 or 64 layers on Roman, because of the substantially higher training cost of the deeper configurations. This difference in the explored depth range may limit the best attainable GCNII performance on the WebKB datasets and should be considered when interpreting those results.

Second, our non-graph MLP baseline is included as a feature-only reference rather than as a competing graph-based method. Its performance therefore provides an estimate of the predictive information available directly from the node representation without neighborhood propagation. The substantial variation in MLP performance across feature spaces, including the increase from 44.30\% with the custom Roman-Empire representation to 76.43\% with RoBERTa, demonstrates that the available node-level information itself can vary considerably across representations.

Finally, all architecture and representation combinations are independently tuned using the procedure described in the main methodology. Consequently, differences in performance reflect the combined effects of the representation, architecture, and the selected optimization configuration. A more restrictive study using matched parameter counts, computational budgets, or jointly fixed hyperparameters could provide additional insight into the interaction between representation and architecture.

\section{Interpreting Feature Sensitivity for Naive GAT}

Naive GAT exhibits the largest contextual gain among the evaluated GNN architectures on Roman-Empire, increasing from 38.96\% Macro F1 with fastText to 52.63\% with RoBERTa, a gain of 13.67 percentage points. This large change should be interpreted alongside the model's absolute performance and fold-to-fold variability. Naive GAT remains the lowest-performing GNN across fastText, BERT, and RoBERTa, while its standard deviation increases from 1.23 percentage points with fastText to 2.67 with BERT and 3.40 with RoBERTa.

Thus, the large contextual gain indicates a strong representation-dependent change for naive GAT, but should not be interpreted as evidence that GAT is the most effective architecture or that contextual representations are uniformly more beneficial for it. In this case, representation sensitivity and absolute model performance are clearly distinct quantities. The result is therefore reported as evidence that architectures can respond differently to changes in node-feature representation, rather than as a ranking of architectures by feature responsiveness.

\section{Per Class Analysis}

\begin{figure}[htbp]
    \centering
    \includegraphics[width=0.5\linewidth]{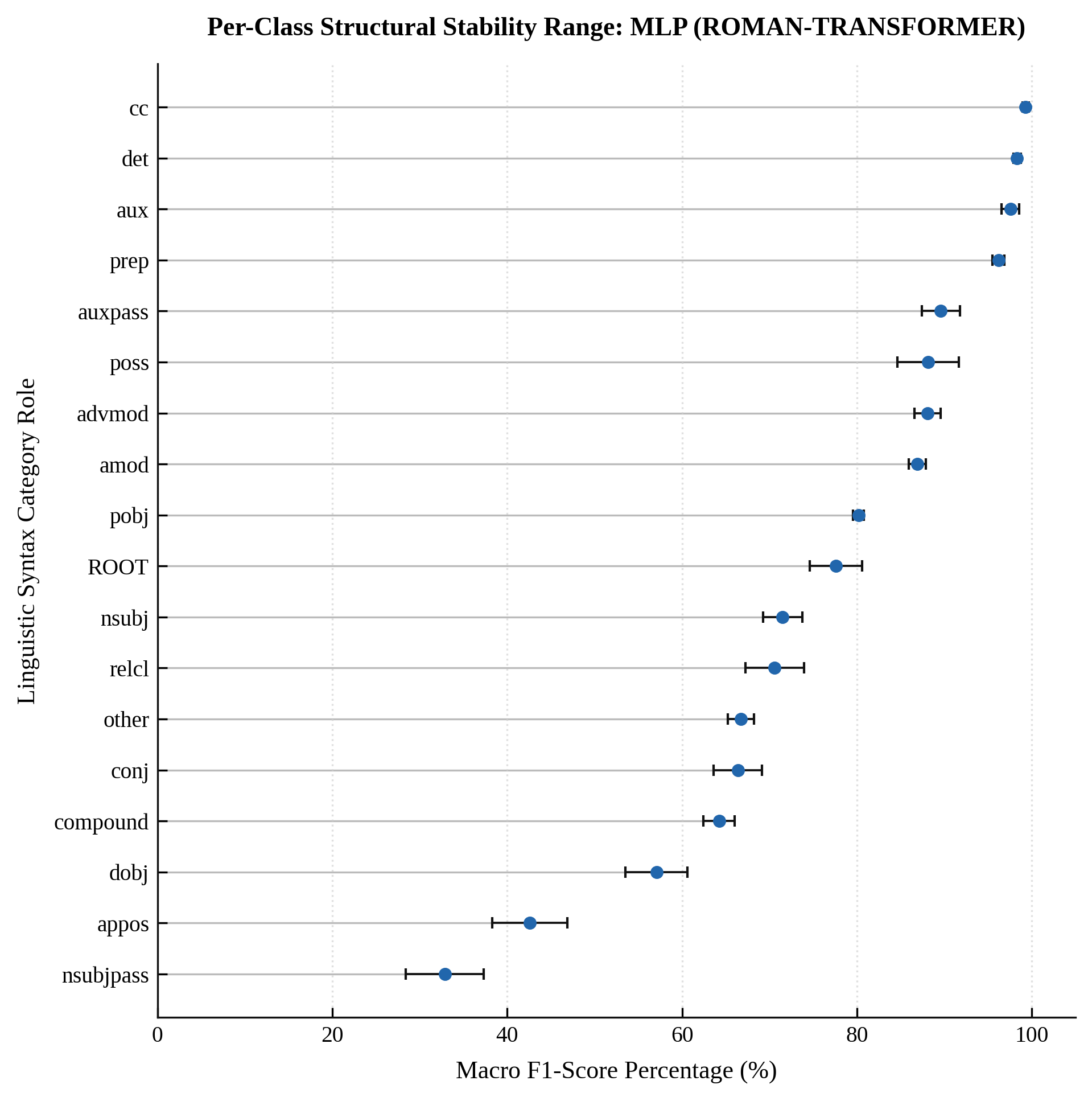}
    \caption{Per-class performance and variability for the feature-only MLP baseline on the Roman-Empire. Performance varies substantially across syntactic categories, indicating that the node representation provides uneven task-relevant information across classes.}
    \label{fig:transformer_dim256_10fold_aggregated_class_breakdown}
\end{figure}

Our per-class analysis of the tuned MLP baseline reveals substantial variation across syntactic categories that is not captured by the aggregate Macro F1 score. As shown in Fig.~\ref{fig:transformer_dim256_10fold_aggregated_class_breakdown}, Class 7 (compound modifiers) and Class 9 (direct objects) achieve F1 scores of approximately 64\% and 58\%, respectively, while Class 14 (passive nominal subjects) reaches only 32\%. These differences suggest that the contextual node representation does not provide equally useful information for all syntactic roles.

\begin{figure}[htbp]
    \centering
    \includegraphics[width=0.5\linewidth]{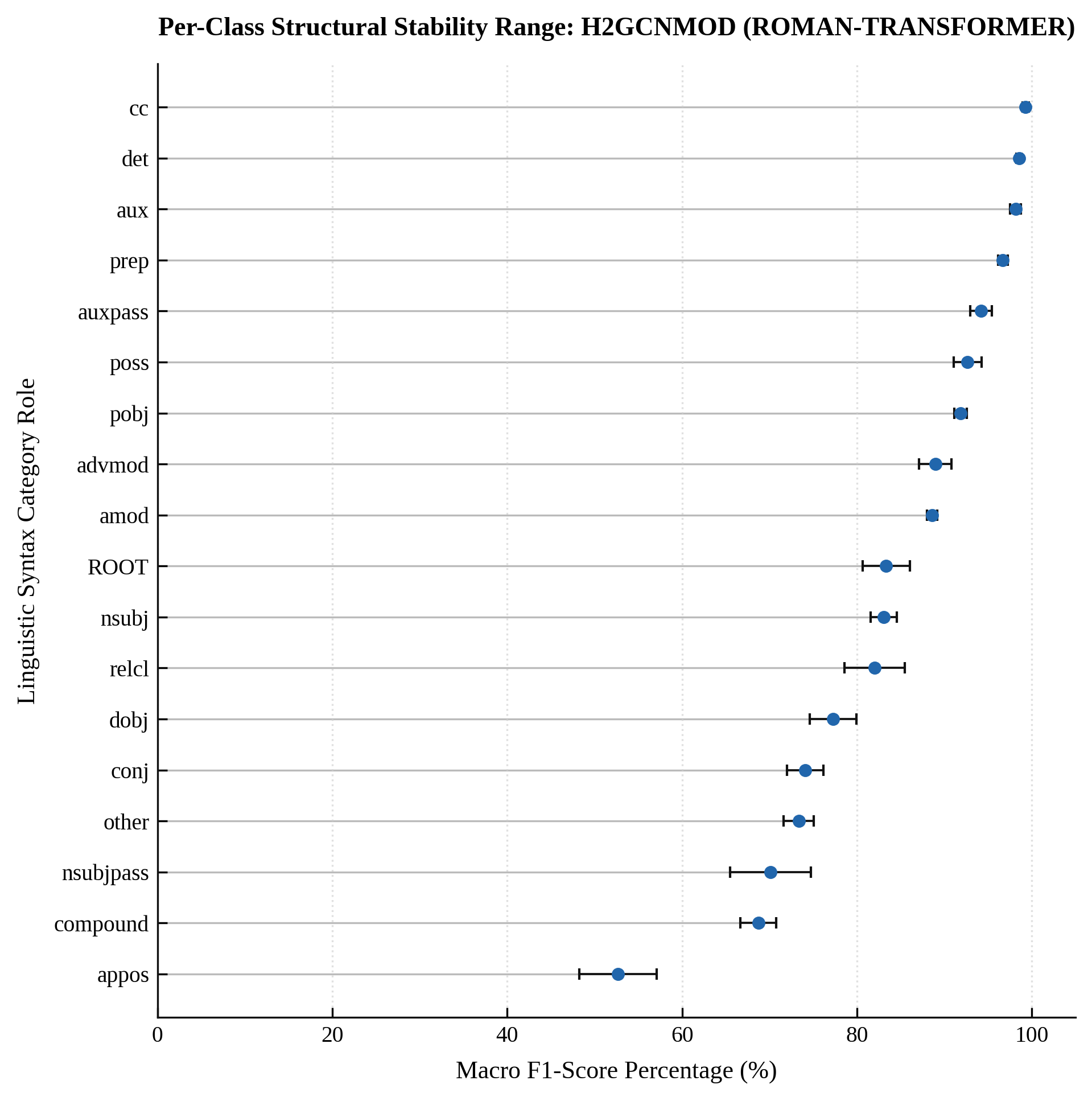}
    \caption{Per-class performance and variability for H2GCN on the Roman-Empire. Compared with the feature-only MLP, graph-based propagation changes performance across syntactic categories, with particularly large differences for selected classes.}
    \label{fig:h2gcnmod_roman}
\end{figure}

When the same node features are processed with H2GCN (Fig.~\ref{fig:h2gcnmod_roman}), Class 14 improves substantially relative to the feature-only MLP. This result suggests that graph-based propagation can provide information that is not fully available from the node representation alone, particularly for categories whose prediction may benefit from relationships among neighboring nodes.

% ==============================================================================
% COMBINED ACCURACY TABLE: ROMAN-EMPIRE (UPDATED WITH NEW ROBERTA RESULTS)
% ==============================================================================
\begin{table}[htbp]
\centering
\caption{Overall Accuracy (OA) and Balanced Accuracy (BA) (\%) across evaluated feature spaces on the Roman-Empire benchmark.}
\label{tab:roman_combined_accuracy}
\small
\begin{tabular}{llcccc}
\toprule
\multirow{2.5}{*}{Model} & \multirow{2.5}{*}{Metric} & \multicolumn{1}{c}{Custom} & \multicolumn{1}{c}{fastText} & \multicolumn{1}{c}{BERT} & \multicolumn{1}{c}{RoBERTa} \\
\cmidrule(lr){3-3} \cmidrule(lr){4-4} \cmidrule(lr){5-5} \cmidrule(lr){6-6}
& & Value & Value & Value & Value \\
\midrule
\multirow{2}{*}{GCN}       & OA & 59.55 $\pm$ 0.81          & 71.66 $\pm$ 0.64          & 70.93 $\pm$ 1.05          & 74.55 $\pm$ 1.93 \\
                           & BA & 51.82 $\pm$ 0.92          & 65.27 $\pm$ 0.89          & 63.92 $\pm$ 1.08          & 68.20 $\pm$ 1.76 \\
\addlinespace
\multirow{2}{*}{GCN-sep}   & OA & \textbf{71.60 $\pm$ 0.78} & 86.26 $\pm$ 0.38          & 88.36 $\pm$ 0.37          & 88.35 $\pm$ 0.59 \\
                           & BA & \textbf{66.77 $\pm$ 1.18} & 83.17 $\pm$ 0.76          & 85.30 $\pm$ 0.91          & 85.52 $\pm$ 0.60 \\
\addlinespace
\multirow{2}{*}{GAT}       & OA & 35.70 $\pm$ 2.15          & 48.83 $\pm$ 1.10          & 54.86 $\pm$ 2.42          & 58.83 $\pm$ 3.08 \\
                           & BA & 27.72 $\pm$ 1.17          & 39.27 $\pm$ 1.18          & 47.79 $\pm$ 2.53          & 51.10 $\pm$ 3.14 \\
\addlinespace
\multirow{2}{*}{GAT-sep}   & OA & 65.68 $\pm$ 0.83          & \textbf{86.81 $\pm$ 0.89} & \textbf{89.95 $\pm$ 0.40} & \textbf{89.58 $\pm$ 0.53} \\
                           & BA & 58.29 $\pm$ 1.28          & \textbf{83.97 $\pm$ 1.24} & \textbf{87.94 $\pm$ 0.63} & \textbf{87.30 $\pm$ 0.60} \\
\addlinespace
\multirow{2}{*}{GraphSAGE} & OA & 65.52 $\pm$ 0.60          & 82.40 $\pm$ 0.73          & 87.78 $\pm$ 0.63          & 87.48 $\pm$ 0.44 \\
                           & BA & 58.17 $\pm$ 1.02          & 78.03 $\pm$ 1.14          & 84.71 $\pm$ 1.00          & 84.25 $\pm$ 0.73 \\
\addlinespace
\multirow{2}{*}{H2GCN}     & OA & 65.87 $\pm$ 0.56          & 80.16 $\pm$ 0.39          & 87.37 $\pm$ 0.53          & 87.04 $\pm$ 0.50 \\
                           & BA & 57.04 $\pm$ 0.70          & 74.62 $\pm$ 0.68          & 82.54 $\pm$ 0.83          & 83.12 $\pm$ 0.61 \\
\addlinespace
\multirow{2}{*}{GCNII}     & OA & 60.23 $\pm$ 0.81          & 76.53 $\pm$ 0.47          & 82.54 $\pm$ 0.57          & 83.78 $\pm$ 0.51 \\
                           & BA & 51.62 $\pm$ 0.79          & 70.25 $\pm$ 0.98          & 76.65 $\pm$ 1.11          & 78.12 $\pm$ 0.50 \\
\midrule
\multirow{2}{*}{MLP (ctrl)}& OA & 51.96 $\pm$ 0.43          & 64.49 $\pm$ 0.75          & 81.57 $\pm$ 0.54          & 81.88 $\pm$ 0.50 \\
                           & BA & 42.57 $\pm$ 0.90          & 54.51 $\pm$ 1.16          & 76.00 $\pm$ 0.82          & 75.91 $\pm$ 0.59 \\
\bottomrule
\end{tabular}
\end{table}

\begin{figure}[htbp]
\centering
\includegraphics[width=\linewidth]{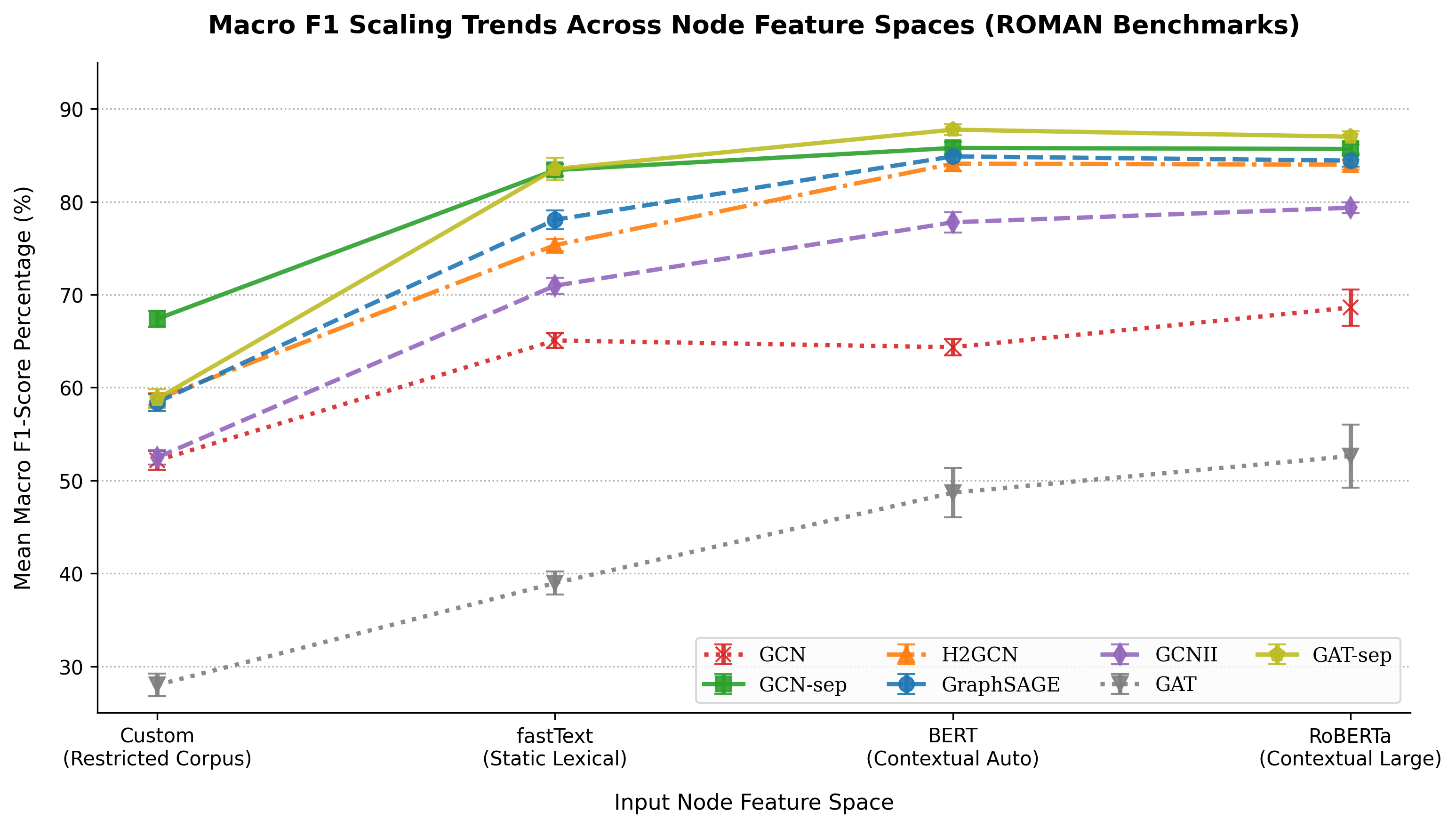}
\caption{Macro F1 scaling trends across restricted corpus, static lexical, and deep contextual node feature spaces on the Roman-Empire benchmark datasets.}
\label{fig:roman_scaling_trends}
\end{figure}

\begin{figure}[htbp]
\centering
\includegraphics[width=\linewidth]{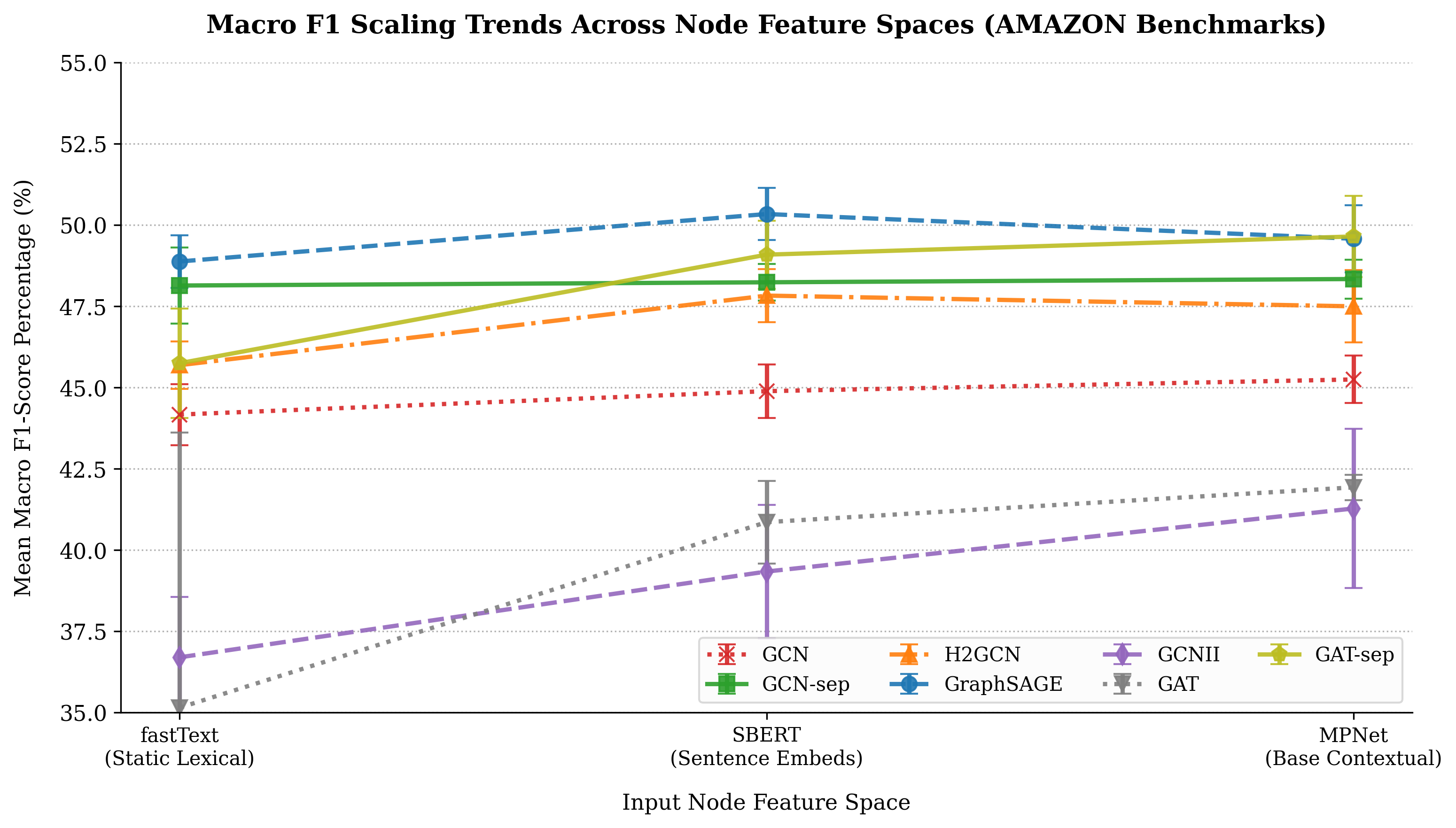}
\caption{Macro F1 scaling trends across static lexical, sentence, and base contextual node feature spaces on the Amazon-Ratings benchmark datasets.}
\label{fig:amazon_scaling_trends}
\end{figure}

\begin{figure}[htbp]
\centering
\includegraphics[width=\linewidth]{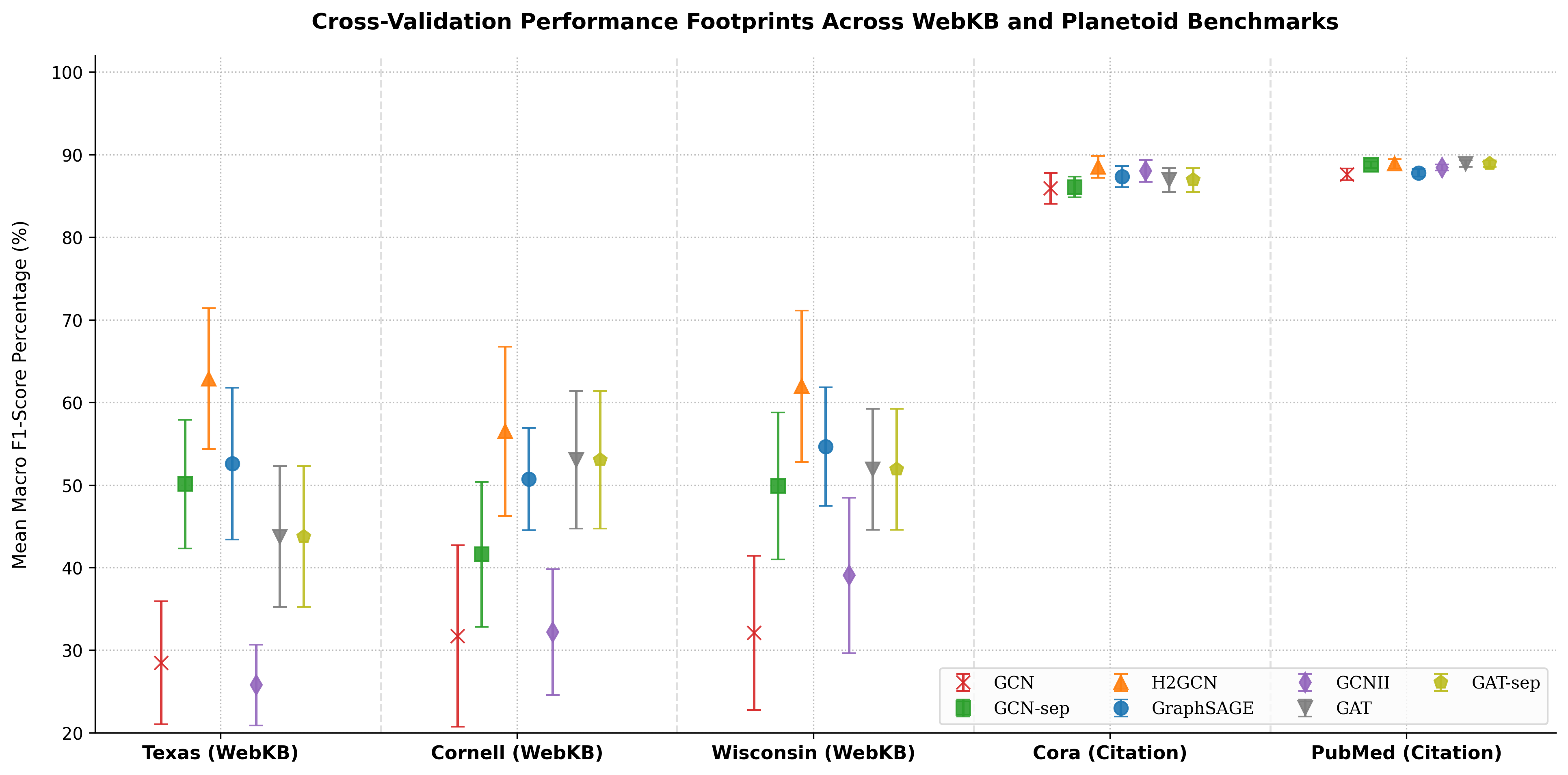}
\caption{Cross-validation performance footprints across WebKB and Planetoid benchmarks, illustrating mean Macro F1-score percentages and associated variance intervals across heterophilic and homophilic graph structures.}
\label{fig:performance_footprints_footprint}
\end{figure}

\begin{figure}[htbp]
    \centering
    \includegraphics[width=\linewidth]{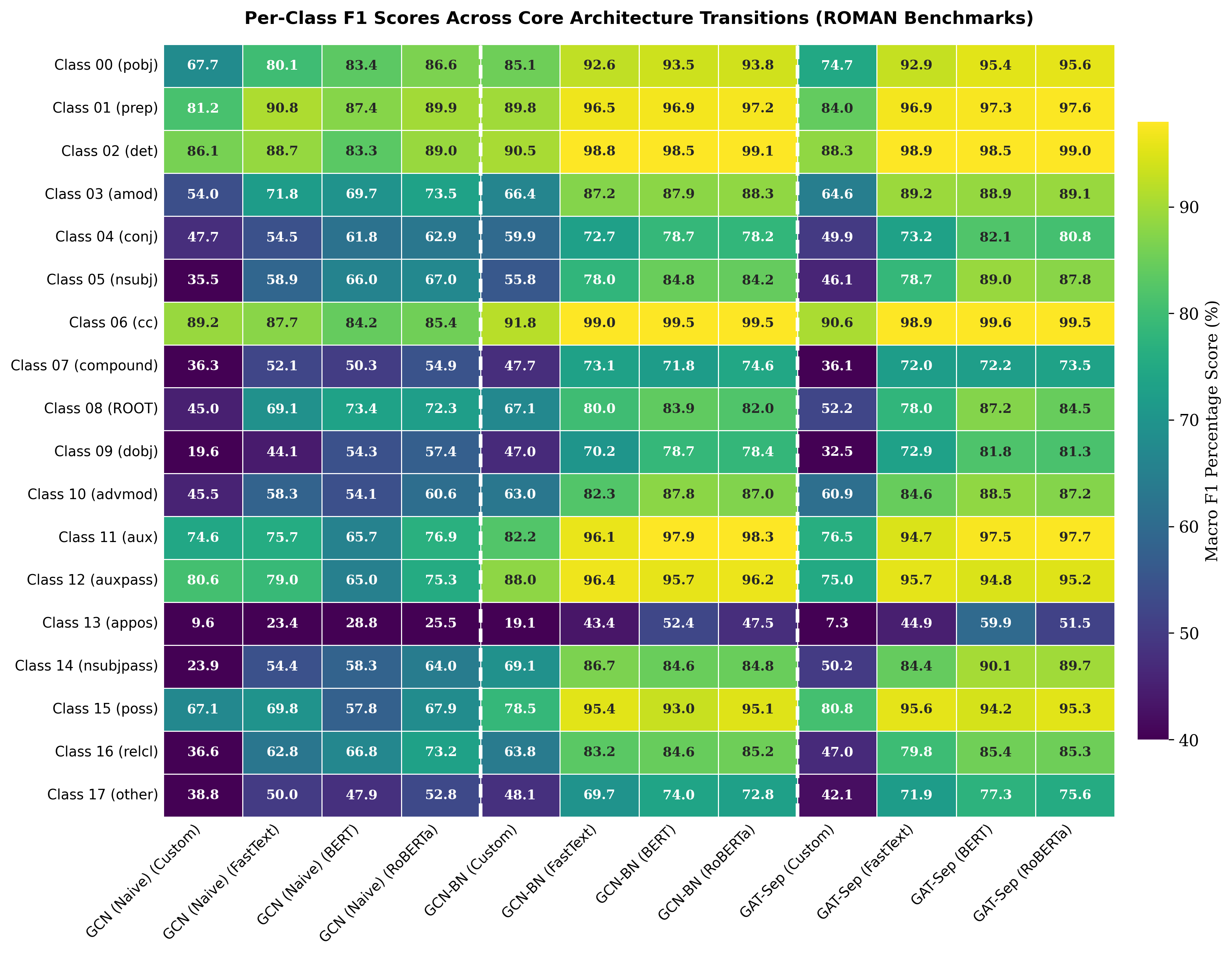}
    \caption{Per-class F1 scores across core network architecture transitions on the Roman-Empire benchmark datasets.}
    \label{fig:appendix_roman_Architecture_story}
\end{figure}

\begin{figure}[htbp]
    \centering
    \includegraphics[width=\linewidth]{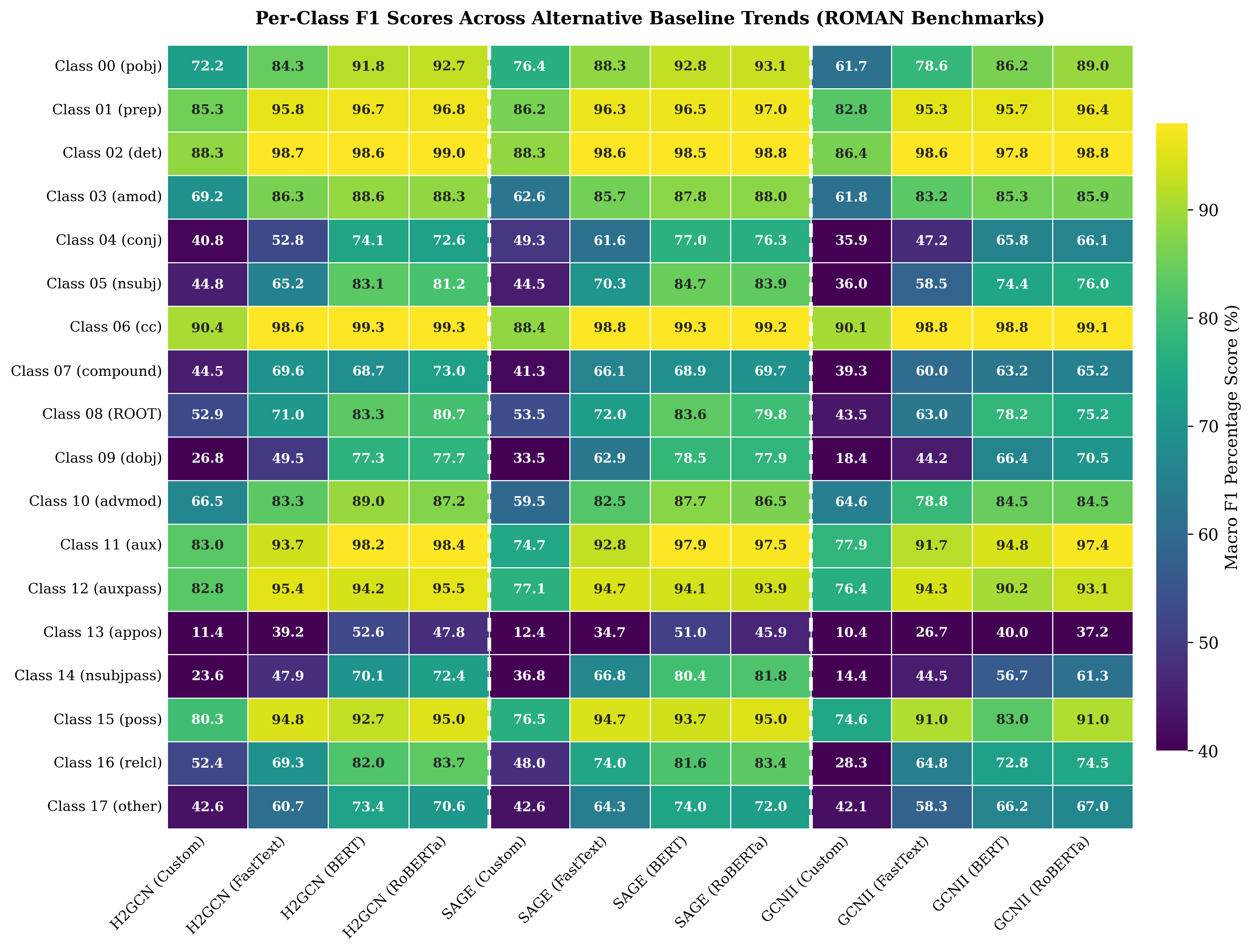}
    \caption{Detailed comparison of per-class F1 scores across feature spaces on the Roman-Empire benchmark, highlighting performance variations across different syntactic dependencies.}
    \label{fig:appendix_roman_feature_story}
\end{figure}

% ==============================================================================
% COMBINED ACCURACY TABLE: AMAZON-RATINGS (FULLY ALIGNED COMPLIANCE)
% ==============================================================================
\begin{table}[htbp]
\centering
\caption{Overall Accuracy (OA) and Balanced Accuracy (BA) (\%) across evaluated feature spaces on the Amazon-Ratings benchmark.}
\label{tab:amazon_combined_accuracy}
\small
\begin{tabular}{llccc}
\toprule
\multirow{2.5}{*}{Model} & \multirow{2.5}{*}{Metric} & \multicolumn{1}{c}{fastText} & \multicolumn{1}{c}{SBERT} & \multicolumn{1}{c}{MPNet} \\
\cmidrule(lr){3-3} \cmidrule(lr){4-4} \cmidrule(lr){5-5}
& & Value & Value & Value \\
\midrule
\multirow{2}{*}{GCN}       & OA & 49.82 $\pm$ 0.86          & 50.55 $\pm$ 0.57          & 50.86 $\pm$ 0.65 \\
                           & BA & 43.55 $\pm$ 0.88          & 44.27 $\pm$ 0.95          & 44.77 $\pm$ 0.64 \\
\addlinespace
\multirow{2}{*}{GCN-sep}   & OA & 53.53 $\pm$ 1.11          & 53.42 $\pm$ 0.42          & 53.84 $\pm$ 0.56 \\
                           & BA & 46.89 $\pm$ 0.81          & 47.70 $\pm$ 0.77          & 47.90 $\pm$ 0.78 \\
\addlinespace
\multirow{2}{*}{GAT}       & OA & 45.47 $\pm$ 4.85          & 49.50 $\pm$ 0.88          & 49.45 $\pm$ 0.63 \\
                           & BA & 35.14 $\pm$ 6.92          & 39.59 $\pm$ 1.14          & 40.94 $\pm$ 0.45 \\
\addlinespace
\multirow{2}{*}{GAT-sep}   & OA & 52.28 $\pm$ 0.98          & 54.84 $\pm$ 0.93          & 55.37 $\pm$ 0.90 \\
                           & BA & 44.63 $\pm$ 1.70          & 47.87 $\pm$ 0.83          & \textbf{48.56 $\pm$ 1.12} \\
\addlinespace
\multirow{2}{*}{GraphSAGE} & OA & \textbf{54.54 $\pm$ 0.60} & \textbf{55.59 $\pm$ 0.63} & \textbf{55.39 $\pm$ 0.57} \\
                           & BA & \textbf{47.66 $\pm$ 0.87} & \textbf{48.80 $\pm$ 0.81} & 47.77 $\pm$ 0.97 \\
\addlinespace
\multirow{2}{*}{H2GCN}     & OA & 52.31 $\pm$ 0.47          & 53.58 $\pm$ 0.63          & 54.59 $\pm$ 0.59 \\
                           & BA & 43.43 $\pm$ 0.57          & 45.60 $\pm$ 0.79          & 44.65 $\pm$ 0.90 \\
\addlinespace
\multirow{2}{*}{GCNII}     & OA & 46.27 $\pm$ 1.34          & 47.44 $\pm$ 1.33          & 48.79 $\pm$ 1.69 \\
                           & BA & 36.32 $\pm$ 1.56          & 38.77 $\pm$ 2.04          & 40.72 $\pm$ 2.33 \\
\midrule
\multirow{2}{*}{MLP (ctrl)}& OA & 47.63 $\pm$ 0.83          & 50.01 $\pm$ 0.38          & 50.86 $\pm$ 0.55 \\
                           & BA & 40.20 $\pm$ 0.77          & 42.67 $\pm$ 0.67          & 42.98 $\pm$ 0.93 \\
\bottomrule
\end{tabular}
\end{table}

\newpage

\section{Hyperparameter Configurations}
\label{sec:appendix_hyperparameters}
% ==============================================================================
% APPENDIX TABLE 1: ROMAN & AMAZON BENCHMARKS
% ==============================================================================

\begin{table*}[p]
\centering
\caption{Optimal hyperparameter configurations for large-scale Roman-Empire and Amazon-Ratings benchmarks across all feature spaces.}
\label{tab:hyperparams_roman_amazon}
\small
\begin{tabular}{lllcccc}
\toprule
\textbf{Category} & \textbf{Dataset Space} & \textbf{Model Name} & \textbf{Layers ($L$)} & \textbf{Hidden Dim ($d$)} & \textbf{LR ($\eta$)} & \textbf{Dropout ($p$)} \\
\midrule
\multirow{32}{*}{\textbf{Roman-Empire}} 
& \multirow{8}{*}{Custom} 
  & GCN       & 5  & 256 & 0.01\hphantom{0} & 0.5 \\
& & GCN-sep   & 5  & 256 & 0.01\hphantom{0} & 0.5 \\
& & GAT (8 heads)     & 1  & 64  & 0.01\hphantom{0} & 0.2 \\
& & GAT-sep (8 heads) & 3  & 64  & 0.005 & 0.2 \\
& & GraphSAGE & 3  & 256 & 0.01\hphantom{0} & 0.5 \\
& & H2GCN     & 2  & 256 & 0.005 & 0.2 \\
& & GCNII     & 32 & 256 & 0.005 & 0.2 \\
& & MLP       & 3  & 128 & 0.01\hphantom{0} & 0.2 \\
\cmidrule{2-7}
& \multirow{8}{*}{fastText} 
  & GCN       & 4  & 256 & 0.01\hphantom{0} & 0.5 \\
& & GCN-sep   & 5  & 128 & 0.01\hphantom{0} & 0.2 \\
& & GAT (8 heads)     & 1  & 64  & 0.01\hphantom{0} & 0.2 \\
& & GAT-sep (8 heads) & 5  & 128 & 0.001 & 0.2 \\
& & GraphSAGE & 3  & 256 & 0.01\hphantom{0} & 0.5 \\
& & H2GCN     & 2  & 256 & 0.005 & 0.2 \\
& & GCNII     & 32 & 256 & 0.005 & 0.2 \\
& & MLP       & 3  & 256 & 0.005 & 0.2 \\
\cmidrule{2-7}
& \multirow{8}{*}{BERT} 
  & GCN       & 3  & 256 & 0.01\hphantom{0} & 0.5 \\
& & GCN-sep   & 3  & 256 & 0.01\hphantom{0} & 0.5 \\
& & GAT (8 heads)     & 1  & 64  & 0.01\hphantom{0} & 0.1 \\
& & GAT-sep (8 heads) & 5  & 128 & 0.005 & 0.2 \\
& & GraphSAGE & 3  & 256 & 0.005 & 0.5 \\
& & H2GCN     & 2  & 256 & 0.005 & 0.5 \\
& & GCNII     & 32 & 256 & 0.005 & 0.2 \\
& & MLP       & 3  & 256 & 0.005 & 0.5 \\
\cmidrule{2-7}
& \multirow{8}{*}{RoBERTa} 
  & GCN       & 5  & 256 & 0.01\hphantom{0} & 0.5 \\
& & GCN-sep   & 3  & 256 & 0.01\hphantom{0} & 0.5 \\
& & GAT (8 heads)     & 1  & 64  & 0.01\hphantom{0} & 0.2 \\
& & GAT-sep (8 heads) & 3  & 64  & 0.005 & 0.2 \\
& & GraphSAGE & 3  & 256 & 0.005 & 0.5 \\
& & H2GCN     & 2  & 256 & 0.001 & 0.2 \\
& & GCNII     & 64 & 128 & 0.005 & 0.2 \\
& & MLP       & 3  & 256 & 0.001 & 0.5 \\
\midrule
\multirow{24}{*}{\textbf{Amazon-Ratings}} 
& \multirow{8}{*}{SBERT} 
  & GCN       & 3  & 256 & 0.005 & 0.1 \\
& & GCN-sep   & 3  & 256 & 0.01\hphantom{0} & 0.5 \\
& & GAT (8 heads)     & 2  & 128 & 0.001 & 0.2 \\
& & GAT-sep (8 heads) & 3  & 128 & 0.001 & 0.2 \\
& & GraphSAGE & 3  & 256 & 0.001 & 0.5 \\
& & H2GCN     & 2  & 256 & 0.005 & 0.2 \\
& & GCNII     & 10 & 256 & 0.005 & 0.2 \\
& & MLP       & 3  & 256 & 0.001 & 0.2 \\
\cmidrule{2-7}
& \multirow{8}{*}{fastText} 
  & GCN       & 3  & 256 & 0.01\hphantom{0} & 0.5 \\
& & GCN-sep   & 3  & 256 & 0.005 & 0.5 \\
& & GAT (8 heads)     & 2  & 64  & 0.01\hphantom{0} & 0.2 \\
& & GAT-sep (8 heads) & 3  & 64  & 0.001 & 0.1 \\
& & GraphSAGE & 3  & 256 & 0.005 & 0.5 \\
& & H2GCN     & 2  & 256 & 0.005 & 0.2 \\
& & GCNII     & 10 & 256 & 0.005 & 0.2 \\
& & MLP       & 3  & 256 & 0.001 & 0.1 \\
\cmidrule{2-7}
& \multirow{8}{*}{MPNet} 
  & GCN       & 4  & 256 & 0.01\hphantom{0} & 0.1 \\
& & GCN-sep   & 3  & 256 & 0.005 & 0.5 \\
& & GAT (8 heads)     & 2  & 128 & 0.001 & 0.2 \\
& & GAT-sep (8 heads) & 3  & 128 & 0.001 & 0.2 \\
& & GraphSAGE & 2  & 256 & 0.001 & 0.2 \\
& & H2GCN     & 2  & 256 & 0.005 & 0.5 \\
& & GCNII     & 10 & 256 & 0.005 & 0.2 \\
& & MLP       & 3  & 256 & 0.001 & 0.5 \\
\bottomrule
\end{tabular}
\end{table*}

% ==============================================================================
% APPENDIX TABLE 2: SMALL-SCALE HETEROPHILIC BENCHMARKS (WEBKB)
% ==============================================================================
\begin{table*}[htbp]
\centering
\caption{Optimal hyperparameter configurations for small-scale WebKB heterophilic benchmarks.}
\label{tab:hyperparams_webkb}
\small
\begin{tabular}{lllcccc}
\toprule
\multirow{2.5}{*}{\textbf{Category}} & \multirow{2.5}{*}{\textbf{Dataset Space}} & \multirow{2.5}{*}{\textbf{Model Name}} & \textbf{Layers} & \textbf{Hidden Dim} & \textbf{Learning Rate} & \textbf{Dropout} \\
& & & \textbf{($L$)} & \textbf{($d$)} & \textbf{($\eta$)} & \textbf{($p$)} \\
\midrule
\multirow{24}{*}{\textbf{Heterophilic}} 
& \multirow{8}{*}{Cornell} 
  & GCN       & 4  & 64  & 0.005 & 0.2 \\
& & GCN-sep   & 4  & 128 & 0.01\hphantom{0} & 0.1 \\
& & GAT (8 heads)     & 1  & 64  & 0.01\hphantom{0} & 0.2 \\
& & GAT-sep (8 heads) & 3  & 64  & 0.01\hphantom{0} & 0.2 \\
& & GraphSAGE & 2  & 128 & 0.001 & 0.2 \\
& & H2GCN     & 2  & 256 & 0.005 & 0.5 \\
& & GCNII     & 10 & 128 & 0.005 & 0.2 \\
& & MLP       & 2  & 256 & 0.001 & 0.5 \\
\cmidrule(lr){2-7}
& \multirow{8}{*}{Texas} 
  & GCN       & 4  & 64  & 0.005 & 0.5 \\
& & GCN-sep   & 5  & 64  & 0.01\hphantom{0} & 0.5 \\
& & GAT (8 heads)     & 2  & 64  & 0.01\hphantom{0} & 0.2 \\
& & GAT-sep (8 heads) & 5  & 64  & 0.005 & 0.1 \\
& & GraphSAGE & 3  & 256 & 0.005 & 0.2 \\
& & H2GCN     & 2  & 256 & 0.005 & 0.2 \\
& & GCNII     & 10 & 128 & 0.005 & 0.2 \\
& & MLP       & 2  & 128 & 0.01\hphantom{0} & 0.1 \\
\cmidrule(lr){2-7}
& \multirow{8}{*}{Wisconsin} 
  & GCN       & 5  & 128 & 0.005 & 0.5 \\
& & GCN-sep   & 3  & 64  & 0.01\hphantom{0} & 0.5 \\
& & GAT (8 heads)     & 1  & 64  & 0.01\hphantom{0} & 0.2 \\
& & GAT-sep (8 heads) & 5  & 128 & 0.005 & 0.2 \\
& & GraphSAGE & 3  & 128 & 0.01\hphantom{0} & 0.5 \\
& & H2GCN     & 2  & 128 & 0.005 & 0.5 \\
& & GCNII     & 10 & 256 & 0.005 & 0.2 \\
& & MLP       & 3  & 256 & 0.01\hphantom{0} & 0.2 \\
\bottomrule
\end{tabular}
\end{table*}

\newpage

% ==============================================================================
% APPENDIX TABLE 3: HOMOPHILIC CITATION CONTROLS
% ==============================================================================
\begin{table*}[htbp]
\centering
\caption{Optimal hyperparameter configurations for homophilic citation-network baseline controls.}
\label{tab:hyperparams_homophilic}
\small
\begin{tabular}{lllcccc}
\toprule
\multirow{2.5}{*}{\textbf{Category}} & \multirow{2.5}{*}{\textbf{Dataset Space}} & \multirow{2.5}{*}{\textbf{Model Name}} & \textbf{Layers} & \textbf{Hidden Dim} & \textbf{Learning Rate} & \textbf{Dropout} \\
& & & \textbf{($L$)} & \textbf{($d$)} & \textbf{($\eta$)} & \textbf{($p$)} \\
\midrule
\multirow{16}{*}{\textbf{Homophilic}} 
& \multirow{8}{*}{Cora} 
  & GCN       & 5  & 128 & 0.01\hphantom{0} & 0.1 \\
& & GCN-sep   & 4  & 256 & 0.001 & 0.5 \\
& & GAT (8 heads)     & 2  & 128 & 0.01\hphantom{0} & 0.2 \\
& & GAT-sep (8 heads) & 3  & 128 & 0.005 & 0.1 \\
& & GraphSAGE & 3  & 256 & 0.001 & 0.2 \\
& & H2GCN     & 2  & 256 & 0.001 & 0.5 \\
& & GCNII     & 10 & 128 & 0.005 & 0.2 \\
& & MLP       & 3  & 64  & 0.001 & 0.5 \\
\cmidrule(lr){2-7}
& \multirow{8}{*}{PubMed} 
  & GCN       & 3  & 256 & 0.005 & 0.5 \\
& & GCN-sep   & 4  & 256 & 0.005 & 0.5 \\
& & GAT (8 heads)     & 2  & 128 & 0.01\hphantom{0} & 0.2 \\
& & GAT-sep (8 heads) & 3  & 64  & 0.01\hphantom{0} & 0.2 \\
& & GraphSAGE & 3  & 64  & 0.01\hphantom{0} & 0.2 \\
& & H2GCN     & 2  & 256 & 0.005 & 0.5 \\
& & GCNII     & 10 & 256 & 0.005 & 0.2 \\
& & MLP       & 2  & 256 & 0.001 & 0.5 \\
\bottomrule
\end{tabular}
\end{table*}

\end{document}